\documentclass[11pt]{article}

\usepackage[preprint]{acl}

\usepackage{times}
\usepackage{latexsym}
\usepackage{amssymb}
\usepackage{amsmath}
\usepackage{hyperref}
\usepackage{url}
\usepackage{multirow}
\usepackage{xcolor}
\usepackage{booktabs}       % professional-quality tables
\usepackage{amsfonts}       % blackboard math symbols
\usepackage{nicefrac}       % compact symbols for 1/2, etc.
\usepackage{graphicx}
\usepackage{microtype}      % microtypography
\usepackage{xcolor}         % colors
\usepackage{cleveref}
\usepackage{algorithmic}
\usepackage{booktabs}   
\usepackage{siunitx}
\usepackage{subfig}
\usepackage{caption}    
\usepackage[T1]{fontenc}
\usepackage{textcomp}   
\usepackage{enumitem}
\usepackage{algorithm}
\crefname{figure}{Figure}{Figure}
\crefname{table}{Table}{Table}
\crefname{equation}{Equation}{Equation}
\crefname{section}{Section}{Section}
\crefname{algorithm}{Algorithm}{Algorithm}
\crefname{appendix}{Appendix}{Appendix}
\crefname{proposition}{Proposition}{Proposition}
\crefname{corollary}{Corollary}{Corollary}

\usepackage[T1]{fontenc}
\usepackage[utf8]{inputenc}

\usepackage[normalem]{ulem}

\usepackage{microtype}

\usepackage{inconsolata}

\usepackage{graphicx}

\title{Trade-R1: Bridging Verifiable Rewards to Stochastic Environments via Process-Level Reasoning Verification}

\author{
  \textbf{Rui Sun\textsuperscript{1}\thanks{Equal contribution.}},
  \textbf{Yifan Sun\textsuperscript{3}$^*$\thanks{Work done during internship at FinStep and StepFun.}},
  \textbf{Sheng Xu\textsuperscript{2}},
  \textbf{Li Zhao\textsuperscript{1}},
  \textbf{Jing Li\textsuperscript{1}},
\\
  \textbf{Daxin Jiang\textsuperscript{1}},
  \textbf{Cheng Hua\textsuperscript{4}$^\ddagger$},
  \textbf{Zuo Bai\textsuperscript{1,2}\thanks{Corresponding authors.}}
\\[0.3em]
  \textsuperscript{1}StepFun,
  \textsuperscript{2}FinStep,
  \textsuperscript{3}Shanghai University of Finance and Economics,
  \textsuperscript{4}Shanghai Jiao Tong University
\\[0.5em]
  \small{\texttt{\{sunrui, zhaoli, futureli, djiang\}@stepfun.com},
    \texttt{yifan.sun@stu.sufe.edu.cn}, \texttt{xusheng@finstep.cn},}
\\
  \small{\texttt{cheng.hua@sjtu.edu.cn}, \texttt{baizuo@\{finstep.cn,stepfun.com\}}}
\vspace{1em}
}

\begin{document}
\maketitle
\begin{abstract}
Reinforcement Learning (RL) has enabled Large Language Models (LLMs) to achieve remarkable reasoning in domains like mathematics and coding, where verifiable rewards provide clear signals. However, extending this paradigm to financial decision is challenged by the market's stochastic nature: rewards are verifiable but inherently noisy, causing standard RL to degenerate into reward hacking. To address this, we propose Trade-R1, a model training framework that bridges verifiable rewards to stochastic environments via process-level reasoning verification. Our key innovation is a verification method that transforms the problem of evaluating reasoning over lengthy financial documents into a structured Retrieval-Augmented Generation (RAG) task. We construct a triangular consistency metric, assessing pairwise alignment between retrieved evidence, reasoning chains, and decisions to serve as a validity filter for noisy market returns. We explore two reward integration strategies: Fixed-effect Semantic Reward (FSR) for stable alignment signals, and Dynamic-effect Semantic Reward (DSR) for coupled magnitude optimization. Experiments on different country asset selection demonstrate that our paradigm reduces reward hacking, with DSR achieving superior cross-market generalization while maintaining the highest reasoning consistency.

\end{abstract}

\section{Introduction}

Large Language Models (LLMs) have evolved from passive information processors to active decision-making agents~\cite{xi2025rise}. In domains like mathematics and coding, Reinforcement Learning (RL) has unlocked remarkable reasoning capabilities~\cite{guo2025deepseek, roziere2023code}, enabled by deterministic, verifiable rewards.

The financial domain presents a fundamental challenge: rewards are verifiable but inherently stochastic. Market returns provide objective feedback, yet their signal-to-noise ratio is low—positive returns may stem from luck rather than sound analysis. Applying standard RL to financial tasks often leads to reward hacking~\cite{gao2023scaling}: models become momentum machines, memorizing historical winners while hallucinating justifications~\cite{Yang2023FinGPTOF}.

We argue that in stochastic environments, the reasoning process must bridge noisy outcomes to valid learning. However, verifying reasoning in finance is non-trivial: financial documents are lengthy, and prompting LLMs to evaluate reasoning over long contexts leads to attention dilution.

To address this, we propose Trade-R1, a framework enabling Reinforcement Learning with Verifiable Rewards (RLVR) in stochastic environments. Our approach comprises two innovations:

\begin{enumerate}
    \item \textbf{Triangular Verification Protocol:} We decouple evidence extraction from logical verification via a RAG approach. We first retrieve relevant evidence via semantic reranking, then evaluate three pairwise consistency scores: Factuality (Evidence $\leftrightarrow$ Reasoning), Deduction (Reasoning $\leftrightarrow$ Decision), and Consistency (Evidence $\leftrightarrow$ Decision). This triangular metric serves as a process-level validity filter.
    
    \item \textbf{Semantic Reward Strategies:} We explore two approaches: (a) Fixed-effect Semantic Reward (FSR), providing a constant alignment incentive; and (b) Dynamic-effect Semantic Reward (DSR), coupling alignment gradients with return magnitude to prevent penalty evasion.
\end{enumerate}

Our contributions are:
\begin{itemize}
    \item \textbf{RLVR for Stochastic Domains:} We identify the lack of process verification as the root cause of reward hacking in financial RL, and propose a RAG paradigm to gate noisy market signals.
    
    \item \textbf{Triangular Consistency Metric:} We design a metric evaluating pairwise consistency between evidence, reasoning, and decisions, ensuring rewards are granted only to logically grounded outputs.
    
    \item \textbf{Cross-Market Validation:} Experiments on A-Share and US markets show that DSR achieves Pareto optimality between returns and reasoning quality, with superior out-of-distribution generalization.
\end{itemize}

\section{Related Work}

\subsection{Reasoning RL: From Verifiable Rewards to Process-Level Supervision}

OpenAI o1 \cite{openai2024o1} and DeepSeek-R1 \cite{deepseek2024} highlight a recent paradigm shift: pairing reinforcement learning with explicit chain-of-thought generation can unlock strong reasoning performance when the environment provides verifiable feedback (e.g., math proofs, unit tests). This line of work motivates extending reasoning RL to decision domains, but also exposes a central gap: in many real-world settings the reward is not directly verifiable, and outcome-only optimization can amplify spurious correlations.

A closely related mitigation is \emph{process-based supervision}, where intermediate reasoning steps receive feedback instead of relying solely on sparse outcomes. Process Reward Models (PRMs) provide dense evaluation signals over reasoning trajectories \cite{lightman2023let}, and prior evidence suggests process supervision improves robustness over outcome-only signals in complex reasoning tasks \cite{uesato2022solving}. Our work is aligned with this direction, but targets a distinct setting: financial markets where rewards are stochastic and delayed, making direct correctness labels unavailable. We therefore treat reasoning itself as a verification bridge, not by annotating every step with human labels, but by verifying whether reasoning is grounded in retrieved evidence and consistent with the final decision.

\subsection{Reward Hacking Under Noisy Objectives}

Reward hacking (specification gaming) is a well-known failure mode in RLHF style alignment, often framed through Goodhart's Law: when a proxy measure becomes the target, it ceases to be a reliable measure \cite{manheim2018categorizing}. In RLHF, this manifests as optimizing a learned reward model that imperfectly approximates the latent human objective \cite{skalse2022defining}. Empirically, over-optimization can yield a characteristic degradation pattern as policies move farther from their reference distribution, consistent with scaling-law analyses of reward model exploitation \cite{gao2023scaling}. Similar over optimization concerns have been observed in direct alignment objectives that implicitly maximize proxy rewards \cite{rafailov2024scaling}.

% In stochastic environments, the problem intensifies: the agent can overfit to noise in the outcome signal, producing policies that appear high-reward on sampled trajectories while lacking stable causal logic. This failure mode is especially acute in finance, where low signal-to-noise ratios and regime shifts make outcome-only training brittle. Our approach directly targets this regime by introducing a process-level validity filter: rewards are granted only when profitability coincides with \emph{logically grounded} reasoning, thereby reducing the degrees of freedom available for hacking noisy returns.

\subsection{Financial Decision Making}

Before LLM-based reasoning agents, financial RL primarily treated trading and portfolio allocation as numerical control from historical price features. Systems such as FinRL \cite{liu2020finrl} and DeepTrader \cite{wang2021deeptrader} optimize risk-adjusted objectives via DRL algorithms (e.g., PPO/A2C/DDPG). While effective at capturing statistical patterns, these methods are typically black-box and vulnerable to overfitting market noise, partly because they do not explicitly model the evidence and logic structure behind decisions.

Recent work explores aligning LLMs to financial objectives via instruction tuning and preference-based learning. FinGPT efforts adapt general LLMs with financial instruction data and alignment techniques \cite{Yang2023FinGPTOF}. To avoid explicit reward-modeling, FinDPO proposes preference optimization tailored to trading signals \cite{iacovides2025findpo}. Complementary to alignment methods, retrieval-augmented financial agents (e.g., FinAgent \cite{zhang2024finagent}, AlphaFin \cite{li2024alphafin}) ground generation in external corpora, improving factuality and traceability.

Most relevant to our work, Trading-R1 \cite{xiao2025tradingr1} represents an early attempt to transplant reasoning RL into trading via curriculum strategies, but it largely remains outcome-driven and thus inherits the core limitation of stochastic returns: optimizing for realized profit alone can encourage policies that exploit volatility rather than learn durable reasoning.

% Trade-R1 addresses this gap by coupling group-based policy optimization with a retrieval-augmented verification protocol, decoupling evidence extraction from logical verification, and enforcing triangular consistency among evidence, reasoning chains, and decisions.

\begin{figure*}
    \centering
    \includegraphics[width=1\textwidth]{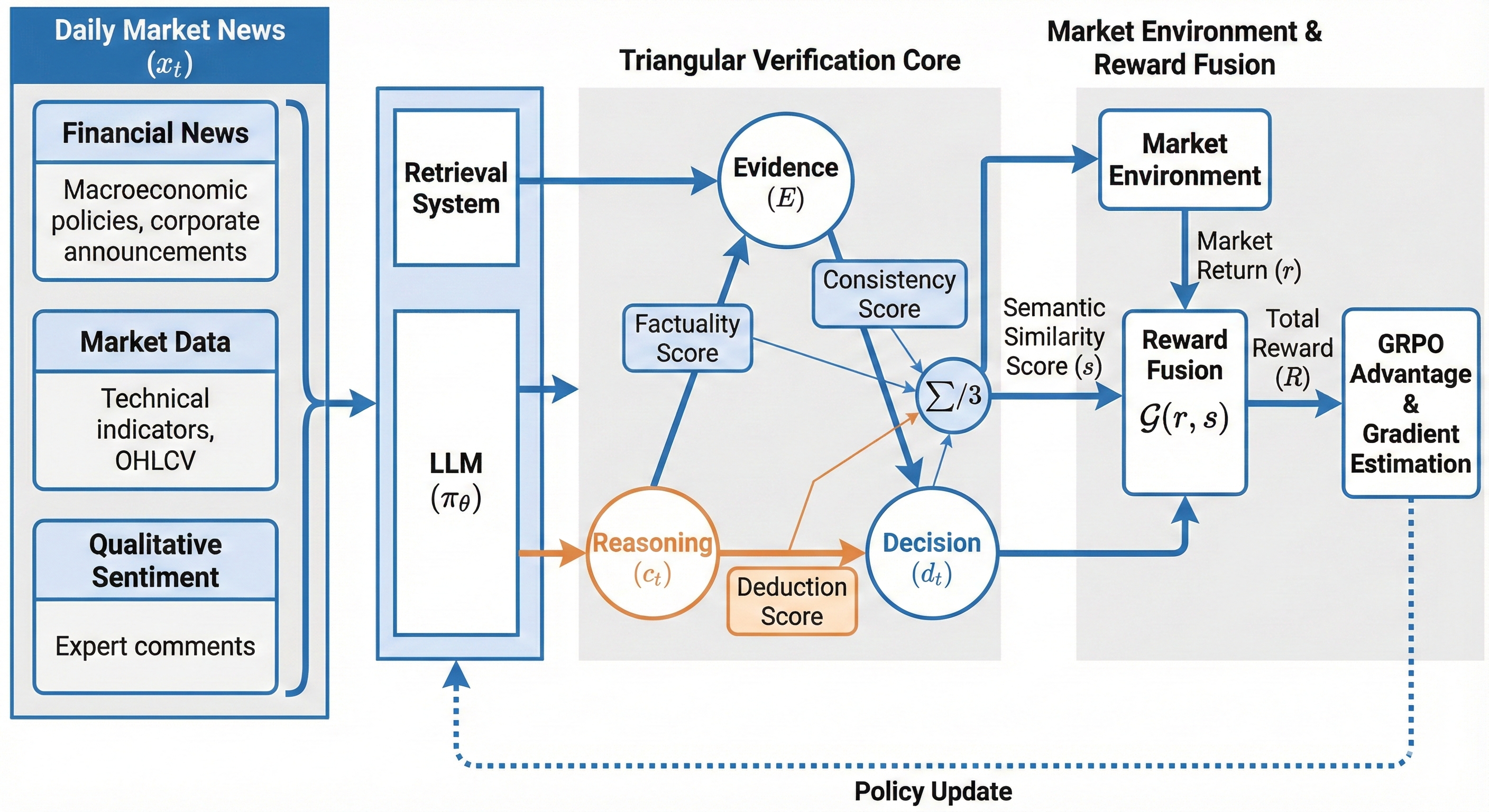}
    \caption{Overview of the Financial LLM Agent Training Architecture Integrating Reasoning Verification and the Asymmetric Semantic Gating (ASG) Mechanism. The system evaluates not only the raw market return ($r$) resulting from the stock selection decision ($d_t$) but also strictly assesses the quality and factual consistency of the model's reasoning process ($c_t$). }
    \label{fig:nav curve}
\end{figure*}

\section{Methodology}

In this section, we first formulate the stock selection problem as a conditional generation task in~\cref{sec:data construction} and~\cref{sec:task formulation}. To mitigate the hacking risk, we introduce the Fixed-effect Semantic Reward (FSR) and Dynamic-effect Semantic Reward (DSR) strategies in~\cref{sec:semantic reward strategy} and present the theoretical analysis in~\cref{sec:theoretical_analysis}. In~\cref{sec:judge}, we detail the verification method used in semantic alignment.

\subsection{Dataset Construction}
\label{sec:data construction}

To capture the multifaceted nature of financial markets, we propose a belief augmented data construction pipeline. The input context $x$ serves as the grounding basis for reasoning and is composed of the following two components:

\textbf{Financial Information Summary.} 
We aggregate financial information from three sources:
\begin{itemize}
    \item \textit{Financial News:} News covering macro-economic policies, corporate announcements, and industry specific regulations.
    \item \textit{Market Data:} Textualized representations of technical indicators (e.g., OHLCV trends) and capital flow data (e.g., smart money movements).
    \item \textit{Qualitative Sentiment:} Transcripts of public interviews with executives and expert market comments.
\end{itemize}

\textbf{Belief Augmentation.}
We augment each daily context with one of $M=15$ distinct investment beliefs, where each belief is a descriptive prompt appended to the raw context to guide the model toward a specific analytical perspective (e.g., dividend detective, blue-chip quality analyst, sector rotation tracker). Mathematically, for a given day $t$ with raw context $C_t$, we generate $M$ distinct input samples $x_{t,k} = (C_t, \textit{Belief}_k)$. The complete list of investment beliefs is provided in~\cref{app:beliefs}.

\subsection{Task Formulation}
\label{sec:task formulation}

We formulate stock selection as a contextual decision problem. At trading day $t$, the model uses previous market context $x_t$ to construct a portfolio $d_t$ aimed at maximizing future excess returns.

\subsubsection{Financial Settings}

We define the asset universe as all A-share stocks, filtering out untradable assets (e.g., ST, suspended, or limit-hit stocks) while retaining newly listed stocks. 

\textbf{Construction \& Weighting.}
During training, the reward is computed for each belief independently based on a market-cap weighted portfolio, enabling the model to learn diverse investment styles. For evaluation, we aggregate outputs from all $M=15$ beliefs via ensemble averaging to obtain a single portfolio, which provides a more stable and interpretable performance metric. Portfolio weights are proportional to market capitalization, and we employ a rolling strategy with 10 overlapping tranches and a 10-day holding period to smooth volatility (details in~\cref{app:finance}).

Directly optimizing for raw market excess return $r$ in stochastic environments leads to reward hacking, where the policy $\pi_\theta$ decouples the decision $d$ from the reasoning $c$. In this case, the model tends to memorize specific assets that have superior performance during the training period, creating a shortcut that bypasses the semantic analysis of the input news. This manifests as spurious prior exploitation, where the model relies on the historical alpha (momentum) of assets rather than the conditional causality provided by the text, leading to poor generalization in out-of-distribution market regimes.

In order to avoid reward hacking, we use semantic-gated reward policy.
Let $\mathcal{D} = \{(x, r)\}$ be a dataset where $x$ represents the market context and $r \in \mathbb{R}$ denotes the scalar market feedback. The policy model $\pi_\theta$ generates a reasoning chain $c$ and a decision $d$. To align with the goal of grounded profitability, our optimization objective maximizes the semantic-gated reward:
\begin{equation}
    J(\theta) = \mathbb{E}_{x \sim \mathcal{D}, y \sim \pi_\theta(\cdot|x)} [\mathcal{G}(r, s)],
\end{equation}
where $\mathcal{G}$ is our proposed asymmetric gating function and $s \in [0,1]$ is the semantic similarity score.

\subsection{Semantic Reward Strategy}
\label{sec:semantic reward strategy}

In this section, we propose two semantic reward strategies: Fixed-effect Semantic Reward (FSR) and Dynamic-effect Semantic Reward (DSR).

\subsubsection{Fixed-effect Semantic Reward}
In this section, we detail the FSR. The reward of FSR is defined as:
\begin{equation}
    \mathcal{G}(r, s) = r +2 \cdot s ,
\label{eq:fsr}
\end{equation}
where $r$ is the market reward and $s$ is the semantic similarity score. In FSR, we set the coefficient to 2 to balance magnitudes. FSR treats semantic alignment as an independent additive term. A key property of FSR is that the incentive to improve reasoning is constant, regardless of the market return ($r$), which provides a stable optimization signal for alignment that is invariant to market volatility.

\subsubsection{Dynamic-effect Semantic Reward}
In this section, we propose DSR, which couples the alignment incentive directly with the magnitude of the market reward, $r$.

We formulate the reward as:
\begin{equation}
\mathcal{G}(r, s) = 
\begin{cases} 
r \cdot (0.5 + s) & \text{if } r > 0 \\
r \cdot (2 - s) & \text{if } r \le 0 
\end{cases}.
\label{eq:reward_def}
\end{equation}

Note that for $r \le 0$, the term simplifies to $r(2-s)$. This formulation effectively scales the magnitude of the financial outcome by the semantic similarity score.

\subsection{Theoretical Analysis}
\label{sec:theoretical_analysis}

To address the challenge of stochastic rewards, we provide a theoretical justification for the DSR strategy. We analyze how coupling semantic validity with market returns inherently improves the optimization landscape by reducing the variance attributable to market noise.

\subsubsection{Problem Formulation: Decomposition of Noisy Returns}
Consistent with standard financial RL formulations, we model the observed market return $r$ as the sum of a latent, deterministic reasoning signal $r^*$ and a stochastic noise term $\xi$:
\begin{equation}
    r = r^* + \xi, \quad \text{where } \xi \sim \mathcal{N}(0, \sigma^2_{\text{noise}}).
\end{equation}
Here, $r^*$ represents the true value justified by the agent's reasoning $y$, while $\xi$ captures market uncertainty. In the Market-only strategy, the gradient estimator $\hat{g}_{\text{mkt}}$ is directly driven by $r$. Consequently, its variance is dominated by the noise variance:
\begin{equation}
    \text{Var}(\hat{g}_{\text{mkt}}) \propto \text{Var}(r) = \text{Var}(r^* + \xi) = \sigma^2_{\text{noise}}.
\end{equation}
High $\sigma^2_{\text{noise}}$ prevents the optimizer from distinguishing between causal reasoning and random reward, leading to reward hacking.

\subsubsection{Variance Analysis of DSR}
\label{subsec:variance}

We analyze the DSR formulation defined in Eq. (3). The modulated reward $\hat{r}_{\text{dsr}}$ is given by:
\begin{align}
    & \mathcal{G}_{\text{dsr}}(r, s) = \alpha(s, r) \cdot r, \\
    & \text{where } \alpha(s,r) = \begin{cases} 0.5 + s & \text{if } r > 0 \\ 2 - s & \text{if } r \le 0 \end{cases}.
\end{align}
Here, $\alpha(s,r)$ acts as a dynamic coefficient. We focus on the positive return regime ($r > 0$), as where reward hacking (overfitting to noise) occurs.

Consider a scenario where the model generates a profitable trade ($r > 0$) driven by market noise ($\xi$ is high) rather than sound reasoning ($r^* \approx 0$). In this case, the semantic similarity score $s$ will be low (e.g., $s \to 0$) because the generated rationale cannot logically verify the serendipitous outcome.
The variance of the DSR gradient estimator in this regime is scaled by the square of the gain coefficient:
\begin{equation}
    \text{Var}(\hat{g}_{\text{dsr}}) \approx \mathbb{E}[(0.5 + s)^2] \cdot \sigma^2_{\text{noise}}.
\end{equation}
As $s \to 0$ (poor reasoning), the gain term $(0.5 + s)^2 \to 0.25$. Comparing this to the standard Market-only:
\begin{equation}
    \text{Var}(\hat{g}_{\text{dsr}}) \approx 0.25 \cdot \text{Var}(\hat{g}_{\text{mkt}}) \ll \text{Var}(\hat{g}_{\text{mkt}}).
\label{eq:var}
\end{equation}
\cref{eq:var} proves that DSR applies strict variance suppression specifically to low-quality, high-noise samples. By dampening the reward signal by 75\% for market volatility, DSR prevents the optimizer from reinforcing spurious correlations.

\subsubsection{Signal Amplification for Valid Reasoning}
Conversely, when the model produces valid reasoning ($s \to 1$) that aligns with a positive market return ($r > 0$), the gain coefficient becomes $(0.5 + 1.0) = 1.5$. This results in a signal amplification effect:
\begin{equation}
    \hat{g}_{\text{dsr}} \approx 1.5 \cdot \hat{g}_{\text{mkt}}.
\end{equation}
Combining these two effects, DSR improves the Signal-to-Noise Ratio (SNR) by simultaneously suppressing the noise variance in low similarity samples and boosting the signal magnitude in high similarity samples:
\begin{equation}
    \text{SNR}_{\text{dsr}} > \text{SNR}_{\text{mkt}}.
\end{equation}
The DSR strategy in the negative domain ($r \le 0$, gain $2-s$) further serves as a penalty regularizer, ensuring that hallucinations resulting in losses are penalized twice.

\subsection{Two-stage Verification}
\label{sec:judge}

% Evaluating similarity score $s$ over long contexts presents significant challenges: LLMs suffer from attention dilution and hallucination. To ensure reproducibility and accuracy, we propose a two-stage verification method in this section, which decouples evidence retrieval from logical verification, transforming the evaluation into a Retrieval-Augmented Generation (RAG) task for the judge.

Evaluating similarity score $s$ over long contexts presents significant challenges. In our task, the average context length approaches 30K tokens. First, existing LLMs exhibit degraded performance on long-context tasks due to attention dilution and increased hallucination rates. Second, evaluation cost becomes a major concern: both monetary expense and latency scale super-linearly with context length, making naive full-context evaluation prohibitively expensive for RL training loops.

To address these challenges, we propose a two-stage verification method that decouples evidence retrieval from logical verification, transforming the evaluation into a Retrieval-Augmented Generation (RAG) task for the judge. This design yields substantial benefits: the average context length for the evaluation task decreases from 30K to 10K tokens, reducing per-step evaluation time by 50\% while simultaneously improving final model performance (see ablation study in~\cref{sec:two stage}).

\subsubsection{Stage 1: Retrieval-Augmented Evidence Extraction}
Instead of feeding the entire noisy input to the judge, we extract a concise evidence context $E$ relevant to the decision $d$. For each selected stock $z \in d$, we first perform hard string matching to locate all mentions in the input $x$. We utilize an embedding model to calculate the semantic similarity between the identified text chunks and the generated output to select top-k relevant chunks. 

\subsubsection{Stage 2: Triangular Similarity Scoring}

Instead of relying on a single scalar score, we decompose the semantic similarity verification into a triangular similarity framework, which evaluates the pairwise consistency between the three critical components: the retrieved evidence ($E$), the reasoning chain ($c$), and the final decision ($d$). We invoke the LLM Judge to compute three distinct similarity scores:

\begin{enumerate}
    \item \textit{Factuality ($\mathcal{S}_{E \leftrightarrow c}$):} Measures whether the reasoning chain $c$ is supported by the facts present in the evidence $E$.
    \item \textit{Deduction ($\mathcal{S}_{c \leftrightarrow d}$):} Evaluates if the final decision $d$ logically follows from the analysis provided in $c$.
    \item \textit{Consistency ($\mathcal{S}_{E \leftrightarrow d}$):} Checks if the decision $d$ aligns with the information contained in the evidence $E$.
\end{enumerate}

The final semantic similarity score $s$ is defined as the arithmetic mean of three components:
\begin{equation}
    s = \frac{1}{3} \left( \mathcal{S}_{E \leftrightarrow c} + \mathcal{S}_{c \leftrightarrow d} + \mathcal{S}_{E \leftrightarrow d} \right) \in [0, 1].
\end{equation}

Above triangular formulation creates a closed logical loop, ensuring high similarity scores are only granted when the output is factually grounded, logically deduced, and consistent with market signals.

We optimize the policy using GRPO, which eliminates the need for a separate value network, thereby reducing memory overhead. For each input $x$, we sample a group of $G$ outputs $\{y_1, \dots, y_G\}$ from the old policy $\pi_{\theta_{old}}$. 

The advantage $A_i$ is computed by normalizing our gated reward within the group:
\begin{equation}
    A_i = \frac{\mathcal{G}(r_i, s_i) - \text{Mean}(\{\mathcal{G}_j\}_{j=1}^G)}{\text{Std}(\{\mathcal{G}_j\}_{j=1}^G) + \epsilon}.
\end{equation}

Group-wise normalization is crucial for financial time-series, as it effectively removes the non-stationary market trend ($\beta$), ensuring the model learns relative asset selection ($\alpha$) conditioned on robust reasoning.

\section{Experiments}
In this section, we evaluate the proposed FSR and DSR strategies on two distinct stock  market: A-share Market (CN) and US stock market. We first detail the datasets, baseline models and hyperparameters in~\cref{sec:setup}. We then compare the portfolio performance and reasoning alignment of FSR and DSR against standard baselines in~\cref{sec:main results}. Finally, we analyze the strategy's effectiveness through ablation studies and robustness checks in~\cref{sec:ablation_asg} and~\cref{sec:two stage}.

\subsection{Experimental Setup}
\label{sec:setup}

\paragraph{Datasets.}
To assess cross-market generalization, we construct two financial reasoning datasets using daily market news briefings from July 2024 to October 2025.
\begin{itemize}
    \item \textit{A-share Market (CN):} Constructed from daily Chinese financial news summaries. The target universe covers all A-share stocks.
    \item \textit{US Stock Market (US):} Constructed from daily English financial news summaries. The target universe covers all stocks listed on major exchanges (e.g., NYSE, NASDAQ), explicitly excluding OTC securities to ensure liquidity.
\end{itemize}
The A-share Market dataset is strictly split by time to prevent look-ahead bias: the Training Set spans from July 2024 to June 2025, while the Test Set covers July 2025 to October 2025. The US Stock Market dataset is utilized exclusively for testing (July 2025 to October 2025) to assess cross-market generalization. For each trading day, we augment the data by pairing the daily news summary with 15 distinct investment style beliefs, generating 15 independent decision samples per day. The market reward $r$ is defined as the 10-day forward excess return relative to the local benchmark (CSI 300).

\paragraph{Implementation Details.}
We employ Qwen3-8B-Instruct as the policy model for both markets, optimized via full-parameter training. The training utilizes the GRPO algorithm with a global batch size of 32 and a learning rate of $1\times 10^{-6}$ (with 10\% warmup). To encourage diverse reasoning exploration, we set the sampling temperature to $1.0$ and generate $G=8$ rollouts per query.
For the semantic alignment verification, we use BGE-M3 as the embedding model to retrieve relevant evidence chunks, and Doubao-seed-1.8 as the judge model to compute the similarity score $s$.

\subsection{Main Results}
\label{sec:main results}
We compare our FSR and DSR strategies with two strategies:
\begin{enumerate}
    \item \textit{Baseline (Qwen3-8B-Instruct):} The pre-trained instruction-following model without any further training, serving as a baseline for the model's intrinsic financial reasoning capabilities.
    \item \textit{Market-Only:} Directly maximizes market reward (10-day excess return to index) without any semantic constraints, representing the standard RL approach in quantitative finance.
\end{enumerate}

We also consider frontier LLMs (Claude-4-Sonnet-thinking, Gemini-2.5-Pro, Deepseek-R1) to compare performance against our FSR and DSR strategies. These frontier models are provided with the same input context as our trained models, ensuring a fair comparison. \Cref{tab:financial_utility} presents a comprehensive quantitative evaluation across two dimensions: Financial Utility (Cumulative Return, Sharpe Ratio, Max Drawdown) and Reasoning Alignment (Semantic Similarity Score, Hallucination Rate).

\begin{figure*}[t]
    \centering
    \begin{minipage}[t]{0.48\textwidth}
        \centering
        \includegraphics[width=\textwidth]{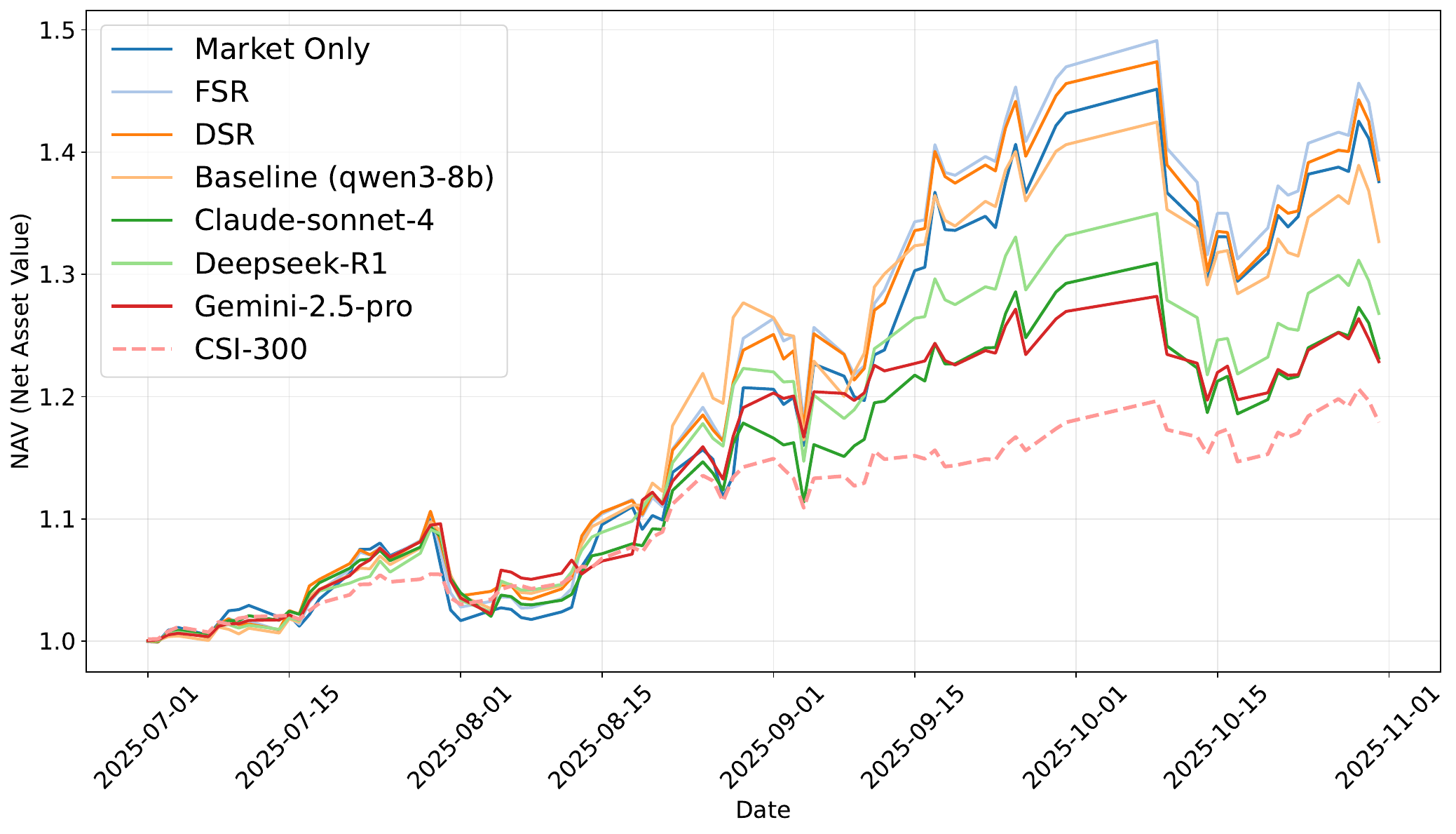}
        \caption{Cumulative net asset value (NAV) curves of different reward strategies on the A-Share market test set (July--October 2025).}
        \label{fig:nav_curve}
    \end{minipage}
    \hfill
    \begin{minipage}[t]{0.48\textwidth}
        \centering
        \includegraphics[width=\textwidth]{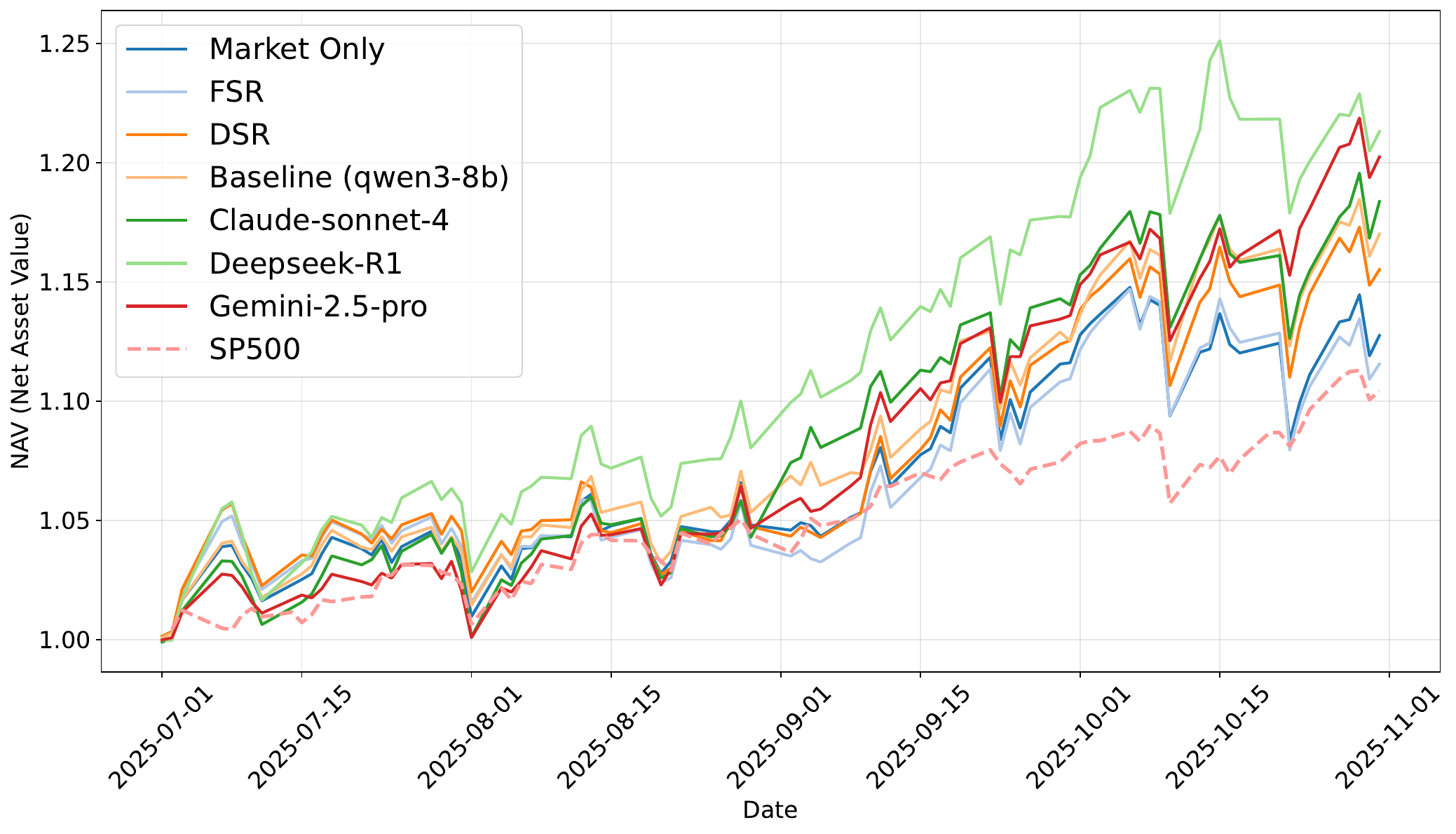}
        \caption{Cumulative net asset value (NAV) curves of different reward strategies on the US market test set (July--October 2025).}
        \label{fig:nav_curve_us}
    \end{minipage}
\end{figure*}

\begin{table*}[t]
\centering
\small
\caption{Comprehensive comparison of financial utility and reasoning quality. The table is organized into four sections: market indices, frontier LLMs (zero-shot), RL baselines, and our proposed methods (FSR and DSR). All experiments for trained models are conducted over 5 random seeds, and the reported data represents the mean values. Detailed standard deviations are provided in the~\cref{app:detailed_results}. All trained models (Market-Only, FSR, DSR) are trained on the A-Share Training Set. Results are reported on the respective test sets for each market. \textbf{Bold}: best. \underline{Underline}: second best. A-Share Index: CSI-300; US Index: S\&P 500.}
\resizebox{2.05\columnwidth}{!}{
\begin{tabular}{l ccc cc ccc cc}
\toprule
\multirow{2}{*}{\textbf{Model}} & \multicolumn{5}{c}{\textbf{A-Share Market}} & \multicolumn{5}{c}{\textbf{US Market}} \\
\cmidrule(lr){2-6} \cmidrule(lr){7-11}
& \textbf{Cum. Ret.}$\uparrow$ & \textbf{Sharpe}$\uparrow$ & \textbf{MDD}$\uparrow$ & \textbf{Sim.}$\uparrow$ & \textbf{Halluc.}$\downarrow$ & \textbf{Cum. Ret.}$\uparrow$ & \textbf{Sharpe}$\uparrow$ & \textbf{MDD}$\uparrow$ & \textbf{Sim.}$\uparrow$ & \textbf{Halluc.}$\downarrow$ \\
\midrule
Index Baseline* & 17.91\% & 3.313 & -4.15\% & - & - & 10.43\% & 2.415 & -2.98\% & - & - \\
\midrule
Claude-4-Sonnet & 23.15\% & 2.575 & \underline{-9.41\%} & 0.5431 & 0.2076 & 18.37\% & 2.573 & \underline{-4.52\%} & 0.5848 & 0.2102 \\
Gemini-2.5-Pro & 22.86\% & 2.780 & \textbf{-6.73\%} & 0.8183 & 0.0555 & \underline{20.24\%} & \textbf{3.114} & \textbf{-3.99\%} & 0.7679 & \textbf{0.0563} \\
Deepseek-R1 & 26.80\% & 2.700 & -9.79\% & 0.5752 & 0.0526 & \textbf{21.30\%} & \underline{2.660} & -5.78\% & 0.4710 & 0.3706 \\
\midrule
Baseline (Qwen3-8B) & 32.60\% & 2.791 & -9.84\% & 0.8970 & 0.0190 & 16.90\% & 2.181 & -5.06\% & \underline{0.7741} & \underline{0.0686} \\
Market Only & 37.62\% & 3.028 & -10.86\% & 0.4369 & 0.2254 & 12.63\% & 1.712 & -5.61\% & 0.6586 & 0.1405 \\
\midrule
FSR (Ours) & \textbf{39.38\%} & \textbf{3.065} & -11.97\% & \underline{0.9560} & \underline{0.0039} & 11.40\% & 1.473 & -5.88\% & 0.7579 & 0.0919 \\
DSR (Ours) & \underline{37.76\%} & \underline{3.036} & -12.04\% & \textbf{0.9744} & \textbf{0.0012} & 15.34\% & 1.951 & -5.40\% & \textbf{0.7768} & 0.0799 \\
\bottomrule
\end{tabular}
}
\label{tab:financial_utility}
\end{table*}

% \begin{table*}[t]
% \centering
% \small
% \resizebox{2.1\columnwidth}{!}{%
% \begin{tabular}{l ccc cc ccc cc}
% \toprule
% \multirow{2}{*}{\textbf{Model}} & \multicolumn{5}{c}{\textbf{A-Share Market Training}} & \multicolumn{5}{c}{\textbf{US Market Training}} \\
% \cmidrule(lr){2-6} \cmidrule(lr){7-11}
% & \textbf{Cum. Ret.} & \textbf{Sharpe} & \textbf{MDD} & \textbf{Sim.} & \textbf{Halluc.} & \textbf{Cum. Ret.} & \textbf{Sharpe} & \textbf{MDD} & \textbf{Sim.} & \textbf{Halluc.} \\
% \midrule
% Market Only & 12.63\% ± 1.23\% & 1.712 ± 0.165 & -5.61\% ± 0.38\% & 0.6586 & 0.1405 & - & - & - & - & - \\
% FSR & 11.40\% ± 2.75\% & 1.473 ± 0.382 & -5.88\% ± 0.54\% & 0.7579 & 0.0919 & - & - & - & - & - \\
% DSR & 15.34\% ± 4.29\% & 1.951317 ± 0.499481 & -5.40\% ± 0.82\% & 0.7767 & 0.0799 & 12.59\% & 1.959 & -4.46\% & 0.7985 & 0.0524 \\
% \bottomrule
% \end{tabular}
% }
% \caption{generalization}
% \label{tab:financial_utility}
% \end{table*}

We analyze the results from three critical perspectives:
\begin{itemize}
    \item \textbf{Failure of Unconstrained RL (Market-Only).}
    The Market-only strategy achieves competitive financial returns on the A-Share market ($37.62\%$ Cum. Ret.) but exhibits catastrophic reasoning degradation: its Similarity Score plummets to $0.4369$, and its Hallucination Rate increases to $0.2254$. This result confirms that without semantic constraints, the model overfits to spurious momentum patterns while generating ungrounded rationales. Furthermore, in-distribution success of Market-only strategy fails to generalize: on the US Market, Market-only yields significantly lower returns with severely degraded reasoning quality compared to DSR.
    \item \textbf{The Limitation of Symmetric Regularization.}
    The FSR strategy achieves the highest financial returns on A-Share ($39.38\%$ Cum. Ret.) while maintaining good alignment (Sim. $0.9560$). However, strong in-distribution performance of FSR fails to transfer effectively on the US Market, FSR produces the lowest returns among all trained models ($11.40\%$), despite maintaining moderate alignment. This reveals that symmetric regularization allows the model to exploit market-specific patterns rather than learning robust, transferable reasoning logic.
    \item \textbf{DSR Achieves Pareto Optimality.}
    Our DSR strategy demonstrates the best trade-off between financial utility and reasoning quality across both markets. On A-Share, while its cumulative return ($37.76\%$) is slightly lower than FSR, DSR achieves: (1) the highest Semantic Similarity Score ($0.9744$), indicating superior reasoning faithfulness; and (2) the lowest Hallucination Rate ($0.0012$), demonstrating factual grounding. Critically, on the out-of-distribution US Market, DSR demonstrates superior cross-market generalization: it achieves higher returns ($15.34\%$) and a better Sharpe ratio ($1.951$) than both Market-Only and FSR, alongside the highest similarity score ($0.7768$). This validates that the Dynamic Effect Semantic Reward effectively couples financial utility with reasoning quality, producing a more robust and trustworthy decision-making model that generalizes across market regimes.
\end{itemize}

\subsection{Ablation Study I: Asymmetric Gating}
\label{sec:ablation_asg}

In this section, we specifically focus on the Dynamic-effect Semantic Reward (DSR), which fundamentally implements an Asymmetric Semantic Gating mechanism. To demonstrate why this asymmetry is crucial, we compare it against a symmetric baseline (Naive Multiply $\mathcal{G}(r,s)=r\cdot s$) and investigate their behaviors under different market conditions.

\begin{figure}
    \centering
    \includegraphics[width=\linewidth]{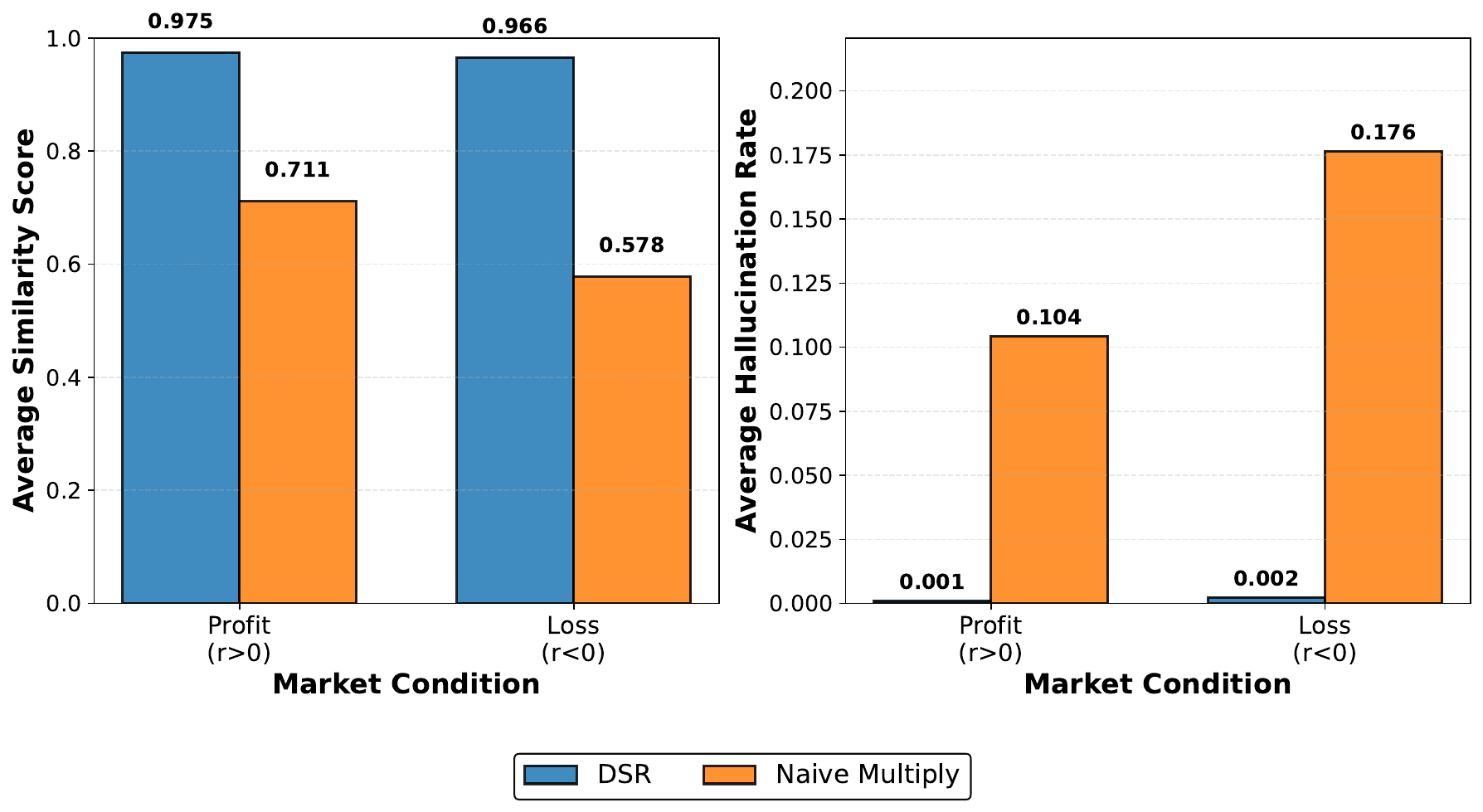}
    \caption{Necessity of Asymmetry. The penalty evasion phenomenon is evident in the symmetric strategy: similarity score drops from $0.711$ (Profit) to $0.578$ (Loss). DSR maintains robustness ($0.966$ in Loss).}
    \label{fig:penalty evasion}
\end{figure}

As illustrated in~\cref{fig:penalty evasion}, the Naive Multiply strategy exhibits a catastrophic drop in semantic similarity scores on loss samples (dropping from $0.711$ to $0.578$) with a higher hallucination rate. The result confirms that without the asymmetric gating mechanism, the model learns to decrease the similarity score in the loss market to mitigate negative rewards. DSR maintains high alignment ($0.966$) even when incurring losses, proving its ability to enforce grounded reasoning in adverse conditions.

\subsection{Ablation Study II: Two-Stage Verification}
\label{sec:two stage}

To assess the effectiveness of our two-stage verification method, we compare our two-stage approach against a simple method that input full context at once, where the LLM receives the raw long context.

\begin{table}[htbp]
\centering
\caption{Necessity of two-stage verification. Comparing the robustness and stability of the verification method. }
\label{tab:Two-Stage}
\resizebox{1\columnwidth}{!}{
\begin{tabular}{lcccc}
\toprule
\textbf{Method} & \textbf{Cum. Ret.} & \textbf{Sharpe} & \textbf{Sim.} & \textbf{Halluc.} \\
\midrule
Simple & 26.45\% ± 0.79\% & 2.912 ± 0.0839 & 0.8068 ± 0.2518 & 0.0266 ± 0.0967\\
Two-Stage & 37.76\% ± 1.65\% & 3.036 ± 0.0809 & 0.9744 ± 0.0602 & 0.0012 ± 0.0154 \\
\bottomrule
\end{tabular}%
}
\end{table}

The results in~\cref{tab:Two-Stage} demonstrate the critical role of the verification method. The Simple method not only yields inferior market excess return ($26.45\%$) but also suffers from high variance in similarity scores ($\pm 0.252$) and a higher hallucination rate. The result demonstrate the limitations of the attention mechanism over long sequences, where the model often fails to attend to the specific evidence segments required for accurate verification. In contrast, our two-stage method ensures precise and stable semantic similarity signals, leading to superior performance across all metrics.
\section{Conclusion}

We presented Trade-R1, a framework bridging verifiable rewards to stochastic environments via process-level reasoning verification. While RL has unlocked remarkable reasoning in deterministic domains like mathematics and coding, applying it to financial decision-making leads to reward hacking, where models become momentum machines that memorize patterns while hallucinating justifications.

Our core insight is that the reasoning process must serve as the bridge to valid learning in stochastic settings. We introduced a Retrieval-Augmented Verification protocol with a Triangular Consistency metric, assessing Factuality, Deduction, and Consistency through pairwise alignment between evidence, reasoning chains, and decisions, enabling accurate process supervision over long financial documents. We explored two semantic reward strategies: Fixed-effect Semantic Reward (FSR) for stable alignment signals, and Dynamic-effect Semantic Reward (DSR) for coupled magnitude optimization that prevents penalty evasion.

% Experiments on A-Share and US markets validate that our paradigm eliminates reward hacking, with DSR achieving Pareto optimality between financial utility and reasoning faithfulness while demonstrating superior cross-market generalization. By treating reasoning quality as a validity filter for noisy outcomes, Trade-R1 enables RLVR in domains where ground-truth labels are unavailable, with potential extensions to medical diagnosis, legal reasoning, and scientific discovery.

\section*{Limitations}

While Trade-R1 demonstrates promising results, we acknowledge several limitations:

\textbf{Temporal Scope.} Our backtesting window is constrained by the LLM's knowledge cutoff to prevent data leakage. Consequently, our method has not been stress-tested across multiple full market cycles (e.g., distinct bull, bear, and high-volatility regimes). The robustness of the triangular consistency metric over longer time horizons remains to be verified.

\textbf{Training Convergence.} Due to computational budget, we terminated training at a pre-defined step rather than waiting for complete convergence. Whether longer training might enable the model to discover subtle strategies to bypass the verification protocol (i.e., ``verifier hacking'') remains an open question.

\textbf{Model and Modality.} We used a fixed model size (Qwen3-8B) and focused exclusively on textual news inputs. The scaling laws of this framework and the integration of multimodal data (e.g., time-series prices, tabular financial statements) are left for future investigation.

\bibliography{custom}

@misc{xi2025rise,
      title={The Rise and Potential of Large Language Model Based Agents: A Survey}, 
      author={Zhiheng Xi and Wenxiang Chen and Xin Guo and Wei He and Yiwen Ding and Boyang Hong and Ming Zhang and Junzhe Wang and Senjie Jin and Enyu Zhou and Rui Zheng and Xiaoran Fan and Xiao Wang and Limao Xiong and Yuhao Zhou and Weiran Wang and Changhao Jiang and Yicheng Zou and Xiangyang Liu and Zhangyue Yin and Shihan Dou and Rongxiang Weng and Wensen Cheng and Qi Zhang and Wenjuan Qin and Yongyan Zheng and Xipeng Qiu and Xuanjing Huang and Tao Gui},
      year={2023},
      eprint={2309.07864},
      archivePrefix={arXiv},
      primaryClass={cs.AI},
      url={https://arxiv.org/abs/2309.07864}, 
}

@article{guo2025deepseek,
   title={DeepSeek-R1 incentivizes reasoning in LLMs through reinforcement learning},
   volume={645},
   ISSN={1476-4687},
   url={http://dx.doi.org/10.1038/s41586-025-09422-z},
   DOI={10.1038/s41586-025-09422-z},
   number={8081},
   journal={Nature},
   publisher={Springer Science and Business Media LLC},
   author={Guo, Daya and Yang, Dejian and Zhang, Haowei and Song, Junxiao and Wang, Peiyi and Zhu, Qihao and Xu, Runxin and Zhang, Ruoyu and Ma, Shirong and Bi, Xiao and Zhang, Xiaokang and Yu, Xingkai and Wu, Yu and Wu, Z. F. and Gou, Zhibin and Shao, Zhihong and Li, Zhuoshu and Gao, Ziyi and Liu, Aixin and Xue, Bing and Wang, Bingxuan and Wu, Bochao and Feng, Bei and Lu, Chengda and Zhao, Chenggang and Deng, Chengqi and Ruan, Chong and Dai, Damai and Chen, Deli and Ji, Dongjie and Li, Erhang and Lin, Fangyun and Dai, Fucong and Luo, Fuli and Hao, Guangbo and Chen, Guanting and Li, Guowei and Zhang, H. and Xu, Hanwei and Ding, Honghui and Gao, Huazuo and Qu, Hui and Li, Hui and Guo, Jianzhong and Li, Jiashi and Chen, Jingchang and Yuan, Jingyang and Tu, Jinhao and Qiu, Junjie and Li, Junlong and Cai, J. L. and Ni, Jiaqi and Liang, Jian and Chen, Jin and Dong, Kai and Hu, Kai and You, Kaichao and Gao, Kaige and Guan, Kang and Huang, Kexin and Yu, Kuai and Wang, Lean and Zhang, Lecong and Zhao, Liang and Wang, Litong and Zhang, Liyue and Xu, Lei and Xia, Leyi and Zhang, Mingchuan and Zhang, Minghua and Tang, Minghui and Zhou, Mingxu and Li, Meng and Wang, Miaojun and Li, Mingming and Tian, Ning and Huang, Panpan and Zhang, Peng and Wang, Qiancheng and Chen, Qinyu and Du, Qiushi and Ge, Ruiqi and Zhang, Ruisong and Pan, Ruizhe and Wang, Runji and Chen, R. J. and Jin, R. L. and Chen, Ruyi and Lu, Shanghao and Zhou, Shangyan and Chen, Shanhuang and Ye, Shengfeng and Wang, Shiyu and Yu, Shuiping and Zhou, Shunfeng and Pan, Shuting and Li, S. S. and Zhou, Shuang and Wu, Shaoqing and Yun, Tao and Pei, Tian and Sun, Tianyu and Wang, T. and Zeng, Wangding and Liu, Wen and Liang, Wenfeng and Gao, Wenjun and Yu, Wenqin and Zhang, Wentao and Xiao, W. L. and An, Wei and Liu, Xiaodong and Wang, Xiaohan and Chen, Xiaokang and Nie, Xiaotao and Cheng, Xin and Liu, Xin and Xie, Xin and Liu, Xingchao and Yang, Xinyu and Li, Xinyuan and Su, Xuecheng and Lin, Xuheng and Li, X. Q. and Jin, Xiangyue and Shen, Xiaojin and Chen, Xiaosha and Sun, Xiaowen and Wang, Xiaoxiang and Song, Xinnan and Zhou, Xinyi and Wang, Xianzu and Shan, Xinxia and Li, Y. K. and Wang, Y. Q. and Wei, Y. X. and Zhang, Yang and Xu, Yanhong and Li, Yao and Zhao, Yao and Sun, Yaofeng and Wang, Yaohui and Yu, Yi and Zhang, Yichao and Shi, Yifan and Xiong, Yiliang and He, Ying and Piao, Yishi and Wang, Yisong and Tan, Yixuan and Ma, Yiyang and Liu, Yiyuan and Guo, Yongqiang and Ou, Yuan and Wang, Yuduan and Gong, Yue and Zou, Yuheng and He, Yujia and Xiong, Yunfan and Luo, Yuxiang and You, Yuxiang and Liu, Yuxuan and Zhou, Yuyang and Zhu, Y. X. and Huang, Yanping and Li, Yaohui and Zheng, Yi and Zhu, Yuchen and Ma, Yunxian and Tang, Ying and Zha, Yukun and Yan, Yuting and Ren, Z. Z. and Ren, Zehui and Sha, Zhangli and Fu, Zhe and Xu, Zhean and Xie, Zhenda and Zhang, Zhengyan and Hao, Zhewen and Ma, Zhicheng and Yan, Zhigang and Wu, Zhiyu and Gu, Zihui and Zhu, Zijia and Liu, Zijun and Li, Zilin and Xie, Ziwei and Song, Ziyang and Pan, Zizheng and Huang, Zhen and Xu, Zhipeng and Zhang, Zhongyu and Zhang, Zhen},
   year={2025},
   month=sep, pages={633–638} }

@article{roziere2023code,
  title={Code llama: Open foundation models for code},
  author={Roziere, Baptiste and Gehring, Jonas and Gloeckle, Fabian and Sootla, Sten and Gat, Itai and Tan, Xiaoqing Ellen and Adi, Yossi and Liu, Jingyu and Sauvestre, Romain and Remez, Tal and others},
  journal={arXiv preprint arXiv:2308.12950},
  year={2023}
}

@misc{Yang2023FinGPTOF,
      title={FinGPT: Open-Source Financial Large Language Models}, 
      author={Hongyang Yang and Xiao-Yang Liu and Christina Dan Wang},
      year={2025},
      eprint={2306.06031},
      archivePrefix={arXiv},
      primaryClass={q-fin.ST},
      url={https://arxiv.org/abs/2306.06031}, 
}

@inproceedings{gao2023scaling,
  title={Scaling laws for reward model overoptimization},
  author={Gao, Leo and Schulman, John and Hilton, Jacob},
  booktitle={International Conference on Machine Learning},
  pages={10835--10866},
  year={2023},
  organization={PMLR}
}

@techreport{openai2024o1,
  title={OpenAI o1 System Card},
  author={OpenAI},
  institution={OpenAI},
  year={2024},
  url={https://openai.com/index/openai-o1-system-card/}
}

@misc{deepseek2024,
  title={DeepSeek-V3 Technical Report},
  author={DeepSeek-AI},
  year={2024},
  url={https://github.com/deepseek-ai/DeepSeek-V3},
  note={Referenced as DeepSeek-R1 in context of reasoning scaling}
}

@misc{xiao2025tradingr1,
      title={Trading-R1: Financial Trading with LLM Reasoning via Reinforcement Learning}, 
      author={Yijia Xiao and Edward Sun and Tong Chen and Fang Wu and Di Luo and Wei Wang},
      year={2025},
      eprint={2509.11420},
      archivePrefix={arXiv},
      primaryClass={q-fin.TR},
      url={https://arxiv.org/abs/2509.11420}, 
}

@inproceedings{wang2021deeptrader,
  title={Deeptrader: a deep reinforcement learning approach for risk-return balanced portfolio management with market conditions embedding},
  author={Wang, Zhicheng and Huang, Biwei and Tu, Shikui and Zhang, Kun and Xu, Lei},
  booktitle={Proceedings of the AAAI conference on artificial intelligence},
  volume={35},
  number={1},
  pages={643--650},
  year={2021}
}

@article{liu2020finrl,
  title={FinRL: A deep reinforcement learning library for automated stock trading in quantitative finance},
  author={Liu, Xiao-Yang and Yang, Hongyang and Chen, Qian and Zhang, Runjia and Yang, Liuqing and Xiao, Bowen and Wang, Christina Dan},
  journal={arXiv preprint arXiv:2011.09607},
  year={2020}
}

@article{skalse2022defining,
  title={Defining and characterizing reward gaming},
  author={Skalse, Joar and Howe, Matthew and Krasheninnikov, Dmitri and Krueger, David},
  journal={Advances in Neural Information Processing Systems},
  volume={35},
  pages={27297--27309},
  year={2022}
}

@inproceedings{manheim2018categorizing,
  title={Categorizing variants of Goodhart's Law},
  author={Manheim, David and Garrabrant, Scott},
  booktitle={arXiv preprint arXiv:1803.04585},
  year={2018}
}

@article{rafailov2024scaling,
  title={Scaling Laws for Reward Model Overoptimization in Direct Alignment Algorithms},
  author={Rafailov, Rafael and Hejna, Joey and Park, Ryan and Chelsea, Finn},
  journal={arXiv preprint arXiv:2406.02900},
  year={2024}
}

@article{lightman2023let,
  title={Let's Verify Step by Step},
  author={Lightman, Hunter and Kosaraju, Vineet and Burda, Yura and Edwards, Harri and Baker, Bowen and Lee, Teddy and Leike, Jan and Schulman, John and Sutskever, Ilya and Cobbe, Karl},
  journal={arXiv preprint arXiv:2305.20050},
  year={2023}
}

@article{uesato2022solving,
  title={Solving math word problems with process-and outcome-based feedback},
  author={Uesato, Jonathan and Kushman, Nate and Kumar, Ramana and Song, Francis and Siegel, Noah and Wang, Lisa and Creswell, Antonia and Layne, Geoffrey and Glaese, DA},
  journal={arXiv preprint arXiv:2211.14275},
  year={2022}
}

@inproceedings{zhang2024finagent,
author = {Zhang, Wentao and Zhao, Lingxuan and Xia, Haochong and Sun, Shuo and Sun, Jiaze and Qin, Molei and Li, Xinyi and Zhao, Yuqing and Zhao, Yilei and Cai, Xinyu and Zheng, Longtao and Wang, Xinrun and An, Bo},
title = {A Multimodal Foundation Agent for Financial Trading: Tool-Augmented, Diversified, and Generalist},
year = {2024},
isbn = {9798400704901},
publisher = {Association for Computing Machinery},
address = {New York, NY, USA},
url = {https://doi.org/10.1145/3637528.3671801},
doi = {10.1145/3637528.3671801},
booktitle = {Proceedings of the 30th ACM SIGKDD Conference on Knowledge Discovery and Data Mining},
pages = {4314–4325},
numpages = {12},
location = {Barcelona, Spain},
series = {KDD '24}
}

@inproceedings{li2024alphafin,
    title = "{A}lpha{F}in: Benchmarking Financial Analysis with Retrieval-Augmented Stock-Chain Framework",
    author = "Li, Xiang  and
      Li, Zhenyu  and
      Shi, Chen  and
      Xu, Yong  and
      Du, Qing  and
      Tan, Mingkui  and
      Huang, Jun",
    editor = "Calzolari, Nicoletta  and
      Kan, Min-Yen  and
      Hoste, Veronique  and
      Lenci, Alessandro  and
      Sakti, Sakriani  and
      Xue, Nianwen",
    booktitle = "Proceedings of the 2024 Joint International Conference on Computational Linguistics, Language Resources and Evaluation (LREC-COLING 2024)",
    month = may,
    year = "2024",
    address = "Torino, Italia",
    publisher = "ELRA and ICCL",
    url = "https://aclanthology.org/2024.lrec-main.69/",
    pages = "773--783"
}

@article{iacovides2025findpo,
  title={FinDPO: Financial Sentiment Analysis for Algorithmic Trading through Preference Optimization of LLMs},
  author={Iacovides, Giorgos and Zhou, Wuyang and Mandic, Danilo},
  journal={arXiv preprint arXiv:2507.18417},
  year={2025}
}

\clearpage
\appendix
% 使用标准的 Section 命令，LaTeX 会自动将其编号为 A
\section{Financial Environment and Implementation Details}
\label{app:finance}

\subsection{Asset Universe and Filtering}
The candidate pool $\mathcal{U}_t$ consists of \textbf{all A-share stocks} listed on the Shanghai and Shenzhen Stock Exchanges. To guarantee realistic execution, we apply a dynamic \textbf{Tradability Filter}:
\begin{enumerate}
    \item \textbf{Risk Status:} Stocks labeled as Special Treatment (ST/ST$^*$) are excluded due to delisting risks.
    \item \textbf{Liquidity Constraints:} Stocks suspended from trading (Halted) or hitting price limits (Limit-Up/Limit-Down) at the opening of day $t$ are excluded.
    \item \textbf{IPO Inclusion:} Newly listed stocks are included to capture potential early-stage alpha.
\end{enumerate}

\subsection{Portfolio Construction and Weighting}
We employ distinct portfolio construction strategies for the training and evaluation phases:

\paragraph{Training Phase (Single-Belief).} 
The model generates a portfolio $d_t$ based on a single reasoning path (belief). The weight $w_{i,t}$ for stock $i \in d_t$ is determined by standard market-capitalization weighting:
\begin{equation}
    w_{i,t} = \frac{\text{Cap}_{i,t}}{\sum_{j \in d_t} \text{Cap}_{j,t}},
\end{equation}
where $\text{Cap}_{i,t}$ denotes the circulating market capitalization of stock $i$ at day $t$.

\paragraph{Backtesting Phase (Vote-Scaled Ensemble).} 
We aggregate outputs from $K=30$ sampled beliefs (votes). The aggregated weight is proportional to both the market capitalization and the number of beliefs voting for the stock ($V_{i,t}$):
\begin{equation}
    w_{i,t}^{agg} = \frac{V_{i,t} \cdot \text{Cap}_{i,t}}{\sum_{j \in \mathcal{U}_t} (V_{j,t} \cdot \text{Cap}_{j,t})},
\end{equation}
To mitigate volatility, we adopt a \textbf{Tranche-based Rolling Strategy}: the total capital is divided into 10 overlapping tranches, with $10\%$ of the capital rebalanced daily based on the new aggregated portfolio signals.

\subsection{Reward Calculation}
The reward $r_t$ represents the \textbf{10-day Forward Excess Return} relative to the benchmark ($\rho_{bm}$), accounting for transaction costs ($\mathcal{C}_{trans} = 0.15\%$ for one-way trading):
\begin{equation}
    r_t = \sum_{i \in d_t} w_{i,t} \left(\rho_{i, t \to t+10} - \rho_{bm, t \to t+10}\right) - 2 \cdot \mathcal{C}_{trans},
\end{equation}
where $\rho_{i, t \to t+10}$ represents the raw return of stock $i$ over the 10-day holding period.

\section{Investment Belief Prompts}
\label{app:beliefs}

We design 15 distinct investment belief prompts to augment the daily market context. Each belief represents a specific analytical perspective, guiding the model to reason from different investment philosophies. \cref{tab:beliefs} provides the complete taxonomy.

\begin{table}[htbp]
\centering
\small
\caption{Taxonomy of 15 investment belief prompts used for context augmentation.}
\resizebox{\columnwidth}{!}{
\begin{tabular}{c l p{5.5cm}}
\toprule
\textbf{ID} & \textbf{Belief Category} & \textbf{Description} \\
\midrule
1 & Dividend Detective & Identifies sustainable high-yield stocks through payout ratio analysis. Tracks dividend history and management commentary. \\
2 & Turnaround Specialist & Seeks distressed companies with new management teams. Analyzes restructuring plans via press releases. \\
3 & Blue-Chip Quality Analyst & Researches companies with wide economic moats. Emphasizes durable competitive advantages and consistent ROIC. \\
4 & Small-Cap Discovery Scout & Finds underfollowed sub-\$500M market cap stocks through local news searches. \\
5 & Sector Rotation Tracker & Times industry moves using economic indicators and relative strength comparisons. \\
6 & Management Quality Assessor & Studies CEO interviews and compensation structures. Searches for insider buying patterns. \\
7 & M\&A Rumor Tracker & Monitors industry consolidation patterns and activist investor moves. \\
8 & Consumer Trends Spotter & Identifies shifting preferences through search trend data and social media buzz. \\
9 & Supply Chain Mapper & Researches supplier/customer relationships. Tracks shipping news and port activity. \\
10 & Cyclical Timing Analyst & Monitors commodity price trends and inventory reports in industrial sectors. \\
11 & Insider Transaction Tracker & Follows Form 4 filings for unusual patterns. Correlates with earnings dates. \\
12 & Special Situations Hunter & Searches for spinoffs, restructurings, and post-bankruptcy equities. \\
13 & Aging Population Thematic & Targets healthcare/services for seniors. Researches demographic shifts. \\
14 & Energy Transition Tracker & Follows utility company CAPEX plans and renewable energy investments. \\
15 & Regulatory Change Scout & Tracks FDA approvals and EPA rulings. Analyzes comment letters for policy clues. \\
\bottomrule
\end{tabular}
}
\label{tab:beliefs}
\end{table}

\clearpage
\section{Detailed Results with Standard Deviations}
\label{app:detailed_results}

This section presents the detailed results with standard deviations omitted in~\cref{tab:financial_utility}.

\begin{table}[htbp]
\centering
\small
\caption{Detailed results on A-Share Market test set (mean ± std).}
\resizebox{\columnwidth}{!}{
\begin{tabular}{l ccc cc}
\toprule
\textbf{Model} & \textbf{Cum. Ret.}$\uparrow$ & \textbf{Sharpe}$\uparrow$ & \textbf{MDD}$\uparrow$ & \textbf{Sim.}$\uparrow$ & \textbf{Halluc.}$\downarrow$ \\
\midrule
Baseline (Qwen3-8B) & 32.60\% ± 2.62\% & 2.791 ± 0.192 & -9.84\% ± 1.19\% & 0.897 ± 0.166 & 0.019 ± 0.087 \\
Market Only & 37.62\% ± 1.77\% & 3.028 ± 0.067 & -10.86\% ± 0.11\% & 0.437 ± 0.354 & 0.225 ± 0.227 \\
\midrule
FSR (Ours) & 39.38\% ± 1.19\% & 3.065 ± 0.066 & -11.97\% ± 0.35\% & 0.956 ± 0.098 & 0.004 ± 0.030 \\
DSR (Ours) & 37.76\% ± 1.65\% & 3.036 ± 0.081 & -12.04\% ± 0.13\% & 0.974 ± 0.060 & 0.001 ± 0.015 \\
\bottomrule
\end{tabular}
}
\label{tab:a_share_details}
\end{table}

\begin{table}[htbp]
\centering
\small
\caption{Detailed results on US Market test set (mean ± std).}
\resizebox{\columnwidth}{!}{
\begin{tabular}{l ccc cc}
\toprule
\textbf{Model} & \textbf{Cum. Ret.}$\uparrow$ & \textbf{Sharpe}$\uparrow$ & \textbf{MDD}$\uparrow$ & \textbf{Sim.}$\uparrow$ & \textbf{Halluc.}$\downarrow$ \\
\midrule
Baseline (Qwen3-8B) & 16.90\% ± 4.01\% & 2.181 ± 0.420 & -5.06\% ± 0.85\% & 0.774 ± 0.284 & 0.069 ± 0.176 \\
Market Only & 12.63\% ± 1.23\% & 1.712 ± 0.165 & -5.61\% ± 0.38\% & 0.659 ± 0.353 & 0.141 ± 0.260 \\
\midrule
FSR (Ours) & 11.40\% ± 2.75\% & 1.473 ± 0.382 & -5.88\% ± 0.54\% & 0.758 ± 0.312 & 0.092 ± 0.211 \\
DSR (Ours) & 15.34\% ± 4.29\% & 1.951 ± 0.499 & -5.40\% ± 0.82\% & 0.777 ± 0.308 & 0.080 ± 0.191 \\
\bottomrule
\end{tabular}
}
\label{tab:us_details}
\end{table}

\section{Case Study}
\label{sec:cases}

% Switch to single column for this section
\onecolumn

In this section, we present a example to illustrate model behaviors.

\noindent\rule{\linewidth}{1pt} % 粗分割线
\textbf{Context \& Instruction} \\
\textbf{Date:} 2025-08-13 \quad \textbf{Belief:} Turnaround specialist
\begin{quote}
    \small\texttt{
    [<Task>\textbackslash n As a professional researcher with specific belief, you need to find opportunities in the market today. You need to submit up to 5 critical analysis suggestions to the investor.\textbackslash n</Task>\textbackslash n\textbackslash n<Trigger\_Time>\textbackslash n 2025-08-13 09:00:00 \textbackslash n</Trigger\_Time>\textbackslash n\textbackslash n <Background\_Information>\textbackslash n \# US Stock Market Macro Analysis Report - August 12, 2025\textbackslash n\textbackslash n\#\# I. Overall Market Overview\textbackslash nOn August 12, 2025, the U.S. equity market posted broad gains, with both the Nasdaq-100 (tracked by QQQ) and S\&P 500 (tracked by SPY) closing higher after an initial intraday dip. Both indices reversed early losses to finish near their daily highs, with QQQ outperforming SPY by 0.17 percentage points. Regular session volume was moderate for QQQ and solid for SPY, with SPY seeing exceptionally high after-hours trading volume.\textbackslash n\textbackslash n\#\# II. Intraday Trading Characteristics\textbackslash n- **Morning Session (9:00-12:00)**: QQQ opened at \$575.15, dipped to its intraday low of \$572.48 during the 10:00-11:00 hour, then began a steady upward trend, closing the morning at \$578.40, a 2.8\% rebound from the intraday low. SPY opened at \$638.29, hit its intraday low of \$636.79 in the 10:00-11:00 hour, then trended higher to close the morning at \$641.23, a 0.7\% increase from the session opening price. The first two hours accounted for 40\% of QQQ’s total regular session volume and 20\% of SPY’s regular session volume.\textbackslash n- **Afternoon Session (12:00-16:00)**: QQQ continued its gradual uptrend, hovering between \$578.83 and \$579.55 through the 13:00-14:00 hour, then tested its daily high of \$580.35 in the final trading hour before closing at \$579.99. SPY maintained slow, steady gains, reaching its daily high of \$642.85 in the 15:00-16:00 hour before closing at \$642.55. The final hour saw the second-highest regular session volume for both ETFs, coinciding with the day’s peak prices.\textbackslash n- **Volume Characteristics**: QQQ’s highest regular session volume was 9,543,191 shares in the 9:00-10:00 hour, tapering off to a low of 2,972,834 shares in the 13:00-14:00 hour before spiking to 8,160,037 shares in the final hour. SPY’s regular session volume peaked at 13,370,476 shares in the first hour, fell to 4,228,266 shares in the 13:00-14:00 hour, then rose to 12,756,729 shares in the final hour. SPY recorded 59,119,904 shares of after-hours volume, far exceeding its regular session hourly averages. Price and volume were largely coordinated, with rising prices paired with declining volume through mid-afternoon, followed by a volume spike as prices hit daily highs.\textbackslash n\textbackslash n\#\# III. Daily Closing Data\textbackslash n- **QQQ** (Nasdaq-100 Index ETF): Closing price \$579.99, up from opening price, daily change +0.84\% (\$4.84), volume 39,859,801.\textbackslash n- **SPY** (S\&P 500 Index ETF): Closing price \$642.55, up from opening price, daily change +0.67\% (\$4.26), volume 117,874,195.\textbackslash n\textbackslash n\#\# IV. Macro Fundamental Summary\textbackslash nThe day’s trading reflected stable buying interest following initial intraday profit-taking, with both major indices staging consistent rebounds from their morning lows to finish near daily highs. Technically, both ETFs closed in positive territory, with QQQ posting a larger percentage gain driven by tech sector strength. Market sentiment was neutral to slightly positive, as evidenced by sustained upward price action after the early dip. Regular session volume was in line with recent 30-day averages for both products, with concentrated buying activity in the final trading hour confirming bullish intraday momentum. SPY’s elevated after-hours volume suggests extended institutional trading activity following the regular session close.\textbackslash n\textbackslash n\#\#\# 1. Corporate Partnerships and M\&A\textbackslash n1. Ultimate Fighting Championship (UFC) announced a new media rights deal with Paramount Skydance Corporation (NASDAQ:PSKY) ahead of confirming plans for a 2026 White House event.\textbackslash n\textbackslash n---\textbackslash n\textbackslash n\#\#\# 2. Corporate Earnings and Sales Performance\textbackslash n1. **Ecopetrol (NYSE:EC)**: Q2 2025 sales hit \$7.061 billion, missing the \$7.240 billion analyst consensus by 2.47\% and falling 15.13\% year-over-year from \$8.320 billion.\textbackslash n2. **Iridex (NASDAQ:IRIX)**: Q2 EPS came in at \$(0.06), missing the \$0.06 consensus estimate by 200\%, while sales rose 7.44\% YoY to \$13.57 million from \$12.63 million.\textbackslash n3. **Siebert Finl (NASDAQ:SIEB)**: Q2 EPS dropped 220\% YoY to \$(0.12) from \$0.10, with sales falling 28.71\% YoY to \$14.874 million from \$20.863 million.\textbackslash n4. **Caribou Biosciences (NASDAQ:CRBU)**: Q2 adjusted EPS of \$(0.35) beat the \$(0.39) consensus by 9.09\%; sales of \$2.667 million beat the \$2.465 million consensus by 8.19\% but fell 23.01\% YoY from \$3.464 million.\textbackslash n5. **Anterix (NASDAQ:ATEX)**: Q1 adjusted EPS of \$(0.48) beat the \$(0.57) consensus by15.79\%; sales of \$1.418 million missed the \$1.508 million consensus by5.97\% and fell7.02\% YoY from \$1.525 million.\textbackslash n6. **Antalpha Platform Holding (NASDAQ:ANTA)**: Q2 EPS of \$0.13 beat the \$0.09 consensus estimate, and sales of \$17.010 million beat the \$16.856 million consensus.\textbackslash n7. **CoreWeave (NASDAQ:CRWV)**: Issued Q3 2025 sales guidance of \$1.260 billion to \$1.300 billion, above the \$1.252 billion analyst consensus; FY2025 sales guidance of \$5.150 billion to \$5.350 billion, above the \$5.039 billion consensus.\textbackslash n8. **Webtoon Entertainment (NASDAQ:WBTN)**: Q2 adjusted EPS of \$0.07 beat the \$(0.15) consensus by145.16\%; sales of \$348.300 million beat the \$341.621 million consensus by1.96\% and rose8.51\% YoY from \$320.972 million.\textbackslash n9. **Eupraxia Pharmaceuticals (NASDAQ:EPRX)**: Q2 EPS of \$(0.26) missed the \$(0.19) consensus by36.84\% and fell52.94\% YoY from \$(0.17).\textbackslash n10. **G. Willi-Food Intl (NASDAQ:WILC)**: Q2 EPS jumped 570\% YoY to \$0.67 from \$0.10; sales rose 22.05\% YoY to \$47.600 million from \$39.000 million.\textbackslash n11. **Erdene Resource Dev (TSX:ERD)**: Reported Q2 EPS of \$(0.01).\textbackslash n\textbackslash n---\textbackslash n\textbackslash n\#\#\# 3. Share Buybacks and Insider Trading\textbackslash n1. **CuriosityStream (NASDAQ:CURI)**: Launched an underwritten secondary public offering by a selling shareholder of its common stock. Underwriters have a 30-day option to purchase an additional 15\% of offered shares at the public offering price minus underwriting discounts and commissions. No offering size was disclosed.\textbackslash n\textbackslash n---\textbackslash n\textbackslash n\#\#\# 4. Regulatory Actions and Compliance Issues\textbackslash nNone\textbackslash n\textbackslash n---\textbackslash n\textbackslash n\#\#\# 5. Corporate Investments and Strategic Initiatives\textbackslash nNone\textbackslash n\textbackslash n---\textbackslash n\textbackslash n\#\#\# 6. Industry Sector Dynamics\textbackslash n1. **Long-Term Stock Performance Highlights**:\textbackslash n - **Spotify (NYSE:SPOT)**: Delivered a 5-year annualized return of 21.54\%, outperforming the market by 7.82\%. A \$1,000 investment made 5 years ago is now worth \$2,742.92, with a current market cap of \$141.98 billion.\textbackslash n - **Eli Lilly (NYSE:LLY)**: Delivered a15-year annualized return of21.42\%, outperforming the market by8.92\%. A \$1,000 investment made15 years ago is now worth \$18,400.86, with a \$577.99 billion market cap.\textbackslash n - **Netflix (NASDAQ:NFLX)**: Delivered a15-year annualized return of32.52\%, outperforming the market by20.01\%. A \$100 investment made15 years ago is now worth \$6,450.30, with a \$520.01 billion market cap.\textbackslash n - **MasTec (NYSE:MTZ)**: Delivered a15-year annualized return of21.24\%, outperforming the market by8.74\%. A \$1,000 investment made15 years ago is now worth \$18,238.38, with a \$14.55 billion market cap.\textbackslash n - **The Hartford Insurance Gr (NYSE:HIG)**: Delivered a5-year annualized return of25.27\%, outperforming the market by11.55\%. A \$100 investment made5 years ago is now worth \$301.38, with a \$36.33 billion market cap.\textbackslash n2. **Analyst Rating Adjustments**:\textbackslash n - Citigroup maintained a Buy rating on Tencent Music (NYSE:TME) and raised its price target from \$23 to \$29.\textbackslash n - Citigroup maintained a Buy rating on Winnebago Industries (NYSE:WGO) and raised its price target from \$33 to \$36.\textbackslash n - Macquarie maintained an Outperform rating on Sea (NYSE:SE) and raised its price target from \$178.2 to \$219.9.\textbackslash n - Barclays maintained an Equal-Weight rating on Fluence Energy (NASDAQ:FLNC) and lowered its price target from \$9 to \$8.\textbackslash n - Barclays maintained an Overweight rating on Tencent Music (NYSE:TME) and raised its price target from \$16 to \$27.\textbackslash n - Leerink Partners maintained an Outperform rating on Cardinal Health (NYSE:CAH) and lowered its price target from \$188 to \$186.\textbackslash n - Leerink Partners maintained an Outperform rating on Biohaven (NYSE:BHVN) and lowered its price target from \$60 to \$50.\textbackslash n\textbackslash n---\textbackslash n\textbackslash n\#\#\# 7. Other Important Information\textbackslash n1. UFC CEO Dana White confirmed plans to host a 2026 July 4 event at the White House to celebrate America's 250th anniversary, which is expected to benefit Paramount Skydance (PSKY) and TKO Group Holdings (NYSE:TKO).\textbackslash n2. **Analyst Sentiment Recaps**:\textbackslash n - 11 analysts covering Camden Prop Trust (NYSE:CPT) issued 1 Bullish,3 Somewhat Bullish,6 Indifferent, and1 Somewhat Bearish ratings over the past quarter.\textbackslash n - 12 analysts covering Walmart (NYSE:WMT) issued3 Bullish and9 Somewhat Bullish ratings over the past quarter, with no bearish or indifferent ratings.\textbackslash n3. Full earnings call transcripts were released for Luminar Technologies (NASDAQ:LAZR) and Rigetti Computing (NASDAQ:RGTI) Q2 FY2025 financial results.\textbackslash n</Background\_Information>\textbackslash n
    \textbackslash n<Belief>\textbackslash nTurnaround specialist. Seeks distressed companies with new management teams. Analyzes restructuring plans via press releases and earnings call transcripts. Focuses on debt reduction progress.\textbackslash n</Belief>\textbackslash n\textbackslash n\textbackslash nYou have exhausted all available research steps and are not allowed to perform further searches or create Actions. \textbackslash nNow please complete the task proposed by the user based on the above research information.\textbackslash n\textbackslash nYour output language is English.\textbackslash nYour output format should be like this:\textbackslash n\textbackslash n<Output>\textbackslash n<signals>\textbackslash n<signal>
    \textbackslash n<has\_opportunity>xxx</has\_opportunity> \# yes or no\textbackslash n<action>xxx</action> \# buy or sell\textbackslash n<symbol\_code>xxx</symbol\_code>
    \textbackslash n<symbol\_name>xxx</symbol\_name 
    \textbackslash n</signal>\textbackslash n<!-- Repeat <signal>...</signal> block for each opportunity you identify, up to 5 signals -->\textbackslash n<!-- Only include signals for genuine opportunities you find in the market -->\textbackslash n</signals>\textbackslash n</Output>]
    }
\end{quote}

\noindent\rule{\linewidth}{0.4pt} % 细分割线
\textbf{Model Output}
\begin{quote}
    \small\texttt{
    \textbf{Reasoning:} 
    <think>\textbackslash nOkay, let's tackle this. The user is a turnaround specialist looking for distressed companies with new management teams, focusing on debt reduction and restructuring plans. They want up to 5 investment signals based on the provided background info.\textbackslash n\textbackslash nFirst, I need to go through the news and market analysis. The macro data shows the US stock market had broad gains, with QQQ and SPY both up. The volume patterns indicate that SPY had a spike in after-hours trading, which might suggest continued institutional interest. But the main focus here is on individual company earnings and other events.\textbackslash n\textbackslash nLooking at the corporate earnings section:\textbackslash n\textbackslash n1. **Ecopetrol (EC)**: Q2 sales missed consensus and had a significant YoY drop. That's a negative, so probably not a good opportunity.\textbackslash n\textbackslash n2. **Iridex (IRIX)**: EPS missed by 200\%, but sales increased YoY. Maybe some issues, but not sure if they're in restructuring.\textbackslash n\textbackslash n3. **Siebert Finl (SIEB)**: Both EPS and sales down YoY. Definitely struggling, but is there new management? The info doesn't mention that. Maybe not.\textbackslash n\textbackslash n4. **Caribou Biosciences (CRBU)**: Adjusted EPS beat by 9\%, sales beat but down YoY. The company might be showing some improvement, but is it distressed? The YoY decline is a red flag. However, beating estimates could indicate potential recovery. Maybe worth considering if there's restructuring mentioned, but the background doesn't say.\textbackslash n\textbackslash n5. **Anterix (ATEX)**: Adjusted EPS beat by 15.79\%, but sales missed and down YoY. Similar to CRBU, mixed results. Not sure about debt reduction.\textbackslash n\textbackslash n6. **Antalpha Platform Holding (ANTA)**: Both EPS and sales beat consensus. Positive, but does this mean they're in a turnaround? Maybe if they were previously underperforming. The background doesn't specify, but beating estimates could be a sign of improvement. However, the user's belief is about distressed companies with new management. If ANTA has new management, this could be a good opportunity. But the info doesn't mention new management, so maybe not.\textbackslash n\textbackslash n7. **CoreWeave (CRWV)**: Sales guidance for Q3 and FY2025 above consensus. This might indicate confidence in their growth, but are they in a distressed state? The user is looking for distressed companies, so unless they're facing issues, this might not fit. However, if their guidance is strong, maybe they're recovering from past problems. Not sure, but possible.\textbackslash n\textbackslash n8. **Webtoon Entertainment (WBTN)**: Adjusted EPS beat by 145\%, sales up YoY. Strong performance, but again, not sure if they're distressed. Maybe they were underperforming before, but the report doesn't mention that. Still, this could be a positive sign.\textbackslash n\textbackslash n9. **G. Willi-Food Intl (WILC)**: EPS jumped 570\% YoY, sales up 22\%. This is a strong positive. If they had new management, this could be a turnaround candidate. The background doesn't mention new management, but the significant improvement might indicate some restructuring or cost-cutting. Could be a good opportunity.\textbackslash n\textbackslash n10. **Eupraxia Pharmaceuticals (EPRX)**: EPS missed and down YoY. Negative, so not a candidate.\textbackslash n\textbackslash n11. **Erdene Resource Dev (ERD)**: EPS of \$(0.01). Not sure, but this is a TSX stock. The user is focused on US market, so maybe not relevant unless there's a specific reason.\textbackslash n\textbackslash nThen, looking at the analyst ratings:\textbackslash n\textbackslash n- Citigroup raised price targets on TME and WGO. But the user is a turnaround specialist, so maybe these are not distressed. TME is Tencent Music, which might be a growth company, not necessarily in distress.\textbackslash n\textbackslash n- Macquarie raised price target on SE (Sea), which is a Singapore company, but listed in NYSE. However, the user's focus is on debt reduction and restructuring. Not sure.\textbackslash n\textbackslash n- Barclays lowered FLNC's target. Not helpful.\textbackslash n\textbackslash n- Leerink Partners lowered CAH and BHVN. Not positive.\textbackslash n\textbackslash nThe industry sector dynamics mention some long-term stock performance, like Netflix and others. But the user is looking for current opportunities, not long-term holders. However, maybe companies like SPOT, LLY, NFLX, MTZ, HIG are performing well, but the user's belief is about distressed companies with new management. Unless they are in a turnaround, these might not fit.\textbackslash n\textbackslash nWait, the user's belief is about companies that are distressed but have new management. So I need to look for companies that are currently underperforming (distressed) but show signs of improvement, possibly due to new management or restructuring.\textbackslash n\textbackslash nLooking back at the earnings:\textbackslash n\textbackslash n- Caribou Biosciences (CRBU) had adjusted EPS beat and sales beat, but YoY sales down 23\%. Maybe they're in a turnaround phase. The user's belief is to analyze restructuring plans via press releases or earnings calls. The background mentions that full transcripts were released for LAZR and RGTI, but not for CRBU. However, the fact that they beat estimates might indicate some positive changes.\textbackslash n\textbackslash n- Anterix (ATEX) had EPS beat but sales missed and down YoY. Maybe they are restructuring, but no info given.\textbackslash n\textbackslash n- G. Willi-Food Intl (WILC) had a huge EPS increase and sales growth. This could be a result of new management or cost-cutting. Since the user is a turnaround specialist, this might be a good candidate. However, the background doesn't mention new management, but the significant improvement could suggest that. Maybe the company is in a turnaround phase.\textbackslash n\textbackslash n- Antalpha (ANTA) beat both EPS and sales. If they were previously struggling, this is a positive. But the background doesn't say they were distressed. However, since the user is looking for opportunities, maybe this is a sign of improvement. But without info on debt reduction or new management, it's uncertain.\textbackslash n\textbackslash n- CoreWeave (CRWV) has strong sales guidance. If they were previously underperforming, this could be a turnaround. But again, no mention of new management or debt issues.\textbackslash n\textbackslash nWait, the user's belief is to focus on companies with new management teams and analyze restructuring plans. The background info includes earnings call transcripts for LAZR and RGTI. Maybe those companies are relevant? But the user is looking for opportunities, not necessarily based on transcripts unless they have restructuring plans.\textbackslash n\textbackslash nLooking at the other sections:\textbackslash n\textbackslash nIn the \"Other Important Information\" section, there's mention of analyst sentiment for Camden Prop Trust (CPT) and Walmart (WMT). For CPT, 11 analysts with mostly bullish ratings. Walmart has all bullish. But the user is a turnaround specialist, so maybe these are not distressed. Unless they were in a turnaround phase.\textbackslash n\textbackslash nBut the main companies that might fit are those with recent EPS improvements, even if they're not explicitly mentioned as distressed. For example, WILC's EPS jumped 570\% YoY. That's a big turnaround. If they have new management, this could be a good buy. However, the background doesn't specify new management, but the user's belief is to look for such cases. Maybe the user can infer that the significant increase is due to a turnaround.\textbackslash n\textbackslash nSimilarly, CRBU beat EPS and sales, but YoY sales down. If they're in a restructuring phase, maybe they can recover. The user's belief includes analyzing restructuring via press releases and earnings calls. Since their Q2 results were released, perhaps they have some restructuring in place.\textbackslash n\textbackslash nAlso, Anterix (ATEX) beat EPS but sales missed. Maybe they're in a turnaround, but not sure.\textbackslash n\textbackslash nAnother point: the user's belief is to focus on debt reduction progress. So, companies that are reducing debt or showing signs of financial recovery.\textbackslash n\textbackslash nLooking at the earnings:\textbackslash n\textbackslash n- WBTN (Webtoon) had a strong EPS beat and sales up YoY. Maybe they are improving, but again, no info on debt.\textbackslash n\textbackslash n- CRWV (CoreWeave) has higher sales guidance. If they were previously underperforming, this could be a sign of restructuring. But again, no direct info.\textbackslash n\textbackslash nWait, the user is a turnaround specialist. So maybe the companies that are currently in the news with positive earnings surprises, even if they have some issues, could be opportunities. For example, CRBU and WBTN beat estimates, which could indicate that they're improving. However, the user needs to check if they are in a distressed state. The problem is that the background doesn't explicitly state that any of these companies are distressed, but maybe their past performance or recent results suggest that.\textbackslash n\textbackslash nAlternatively, maybe companies with high debt or negative earnings in the past but now showing improvement. For example, Siebert Finl (SIEB) had a 220\% drop in EPS and 28.71\% sales decline. That's a big problem. However, the user is looking for new management teams. If the company has new management, maybe they are restructuring. But the background doesn't mention that. So it's unclear.\textbackslash n\textbackslash nAnother angle: the UFC deal with PSKY and the White House event could be a positive for PSKY and TKO. But the user's belief is about distressed companies, so unless PSKY is in trouble, this might not be relevant. However, the user is a turnaround specialist, so maybe if PSKY is in a restructuring phase, this partnership could help. But the background doesn't mention any distress for PSKY, just a new media rights deal.\textbackslash n\textbackslash nHmm. The user might need to consider companies that have beaten earnings despite negative YoY, which could indicate a turnaround. For example, WILC's EPS up 570\% YoY. If they have new management, that's a good signal. But without explicit mention of new management, it's a bit of a stretch. However, the user's belief allows for analysis via press releases and earnings call transcripts. If the earnings call transcripts for WILC were released, but the background only mentions LAZR and RGTI. So maybe not.\textbackslash n\textbackslash nWait, the background says that full earnings call transcripts were released for Luminar Technologies (LAZR) and Rigetti Computing (RGTI). The user's belief is to analyze restructuring plans via these transcripts. So if those companies have restructuring plans, that could be opportunities. But the background doesn't mention any specific restructuring details for LAZR or RGTI. It just says the transcripts were released. So maybe not enough info.\textbackslash n\textbackslash nAlternatively, the user might focus on companies that are showing strong performance in their earnings, even if they have past issues. For example, CRBU had adjusted EPS beat and sales beat, but YoY sales down. If they're in a phase of restructuring, maybe they can recover. The user's belief includes analyzing restructuring plans, so perhaps the fact that they beat estimates is a sign that restructuring is working.\textbackslash n\textbackslash nSimilarly, Anterix (ATEX) beat EPS but sales missed. Maybe they're in a turnaround phase. However, without explicit mention of debt reduction or new management, it's uncertain.\textbackslash n\textbackslash nWait, the user is allowed to use the information given. The background mentions that the user is a turnaround specialist, so they need to look for companies that are currently in a turnaround, possibly with new management. The key is to find companies that have recent positive earnings surprises, possibly with restructuring.\textbackslash n\textbackslash nLooking again:\textbackslash n\textbackslash n- Webtoon (WBTN): adjusted EPS of \$0.07 beat the \$(0.15) consensus by 145\%. That's a big positive. Sales also beat and up YoY. This could indicate that the company is recovering. Maybe they have new management or are restructuring. The user might consider this as an opportunity.\textbackslash n\textbackslash n- G. Willi-Food (WILC): EPS up 570\% YoY, sales up 22\%. This is a strong turnaround. If the company has new management or is restructuring, this could be a good buy. The user's belief is to seek such companies. Even if not explicitly stated, the large EPS improvement might suggest that.\textbackslash n\textbackslash n- Caribou Biosciences (CRBU): adjusted EPS beat by 9\%, sales beat by 8\% but down YoY. Maybe they have a new management team that's improving performance, even though YoY is down. The user could look into this.\textbackslash n\textbackslash n- Antalpha (ANTA): beat both EPS and sales. If they were previously underperforming, this could be a turnaround. The user might need to check if they have new management, but the background doesn't say. However, the fact that they beat the consensus could be a signal.\textbackslash n\textbackslash n- CoreWeave (CRWV): sales guidance above consensus. If they were previously struggling, this could indicate restructuring. But again, no direct info on debt or management changes.\textbackslash n\textbackslash nSo, considering the above, the user might select companies that have beaten earnings, even if some metrics are down. Let's list the ones that beat EPS:\textbackslash n\textbackslash n- CRBU: beat adjusted EPS by 9.09\%.\textbackslash n- ATEX: beat adjusted EPS by 15.79\%.\textbackslash n- ANTA: beat EPS and sales.\textbackslash n- WBTN: beat EPS by 145\%.\textbackslash n- WILC: EPS up 570\% YoY.\textbackslash n\textbackslash nBut the user's belief is about distressed companies with new management. So, perhaps the companies that are currently underperforming but have positive earnings surprises. However, the background doesn't specify which of these are distressed. But the user is a turnaround specialist, so they might look for companies that have shown improvement in their recent results, even if not explicitly stated as distressed.\textbackslash n\textbackslash nAlternatively, the user could consider companies with high after-hours volume, like SPY. But SPY is an ETF, not an individual company. The user is looking for individual stocks.\textbackslash n\textbackslash nAnother angle: the analyst ratings. For example, Citigroup raised price targets on TME and WGO. But TME is Tencent Music, which is a NYSE stock. However, the user's focus is on distressed companies. Maybe these are not distressed, but the user might still consider them if they have positive momentum.\textbackslash n\textbackslash nBut the user's belief is about debt reduction and restructuring. So, perhaps the best candidates are those that have shown strong EPS beats and positive sales trends, even if some YoY data is down. For example:\textbackslash n\textbackslash n1. **Webtoon Entertainment (WBTN)**: Beat EPS by 145\% and sales up YoY. This could indicate a successful turnaround. The user might want to buy here.\textbackslash n\textbackslash n2. **G. Willi-Food Intl (WILC)**: Huge EPS increase and sales growth. Could be a turnaround, so buy.\textbackslash n\textbackslash n3. **Caribou Biosciences (CRBU)**: Beat EPS and sales, though sales down YoY. Maybe they are in restructuring, so buy.\textbackslash n\textbackslash n4. **Antalpha (ANTA)**: Beat both EPS and sales. Positive signs, so buy.\textbackslash n\textbackslash n5. **CoreWeave (CRWV)**: Sales guidance above consensus, which might indicate restructuring efforts. Buy.\textbackslash n\textbackslash nBut the user's belief is to look for companies with new management teams and restructuring plans. Since the background doesn't explicitly mention new management for these companies, but the user is allowed to analyze based on the info given. The fact that they beat estimates could be a sign of restructuring. So, perhaps the user can suggest buying these stocks as they show positive momentum and potential for further improvement.\textbackslash n\textbackslash nAlternatively, maybe the user should focus on the companies that have the most significant improvements, like WBTN and WILC. Also, the analyst ratings for TME and WGO are positive, but again, not sure if they are distressed.\textbackslash n\textbackslash nAnother point: the user's belief includes analyzing debt reduction progress. If a company is reducing debt, that's a good sign. However, the background doesn't mention debt details for any of the companies. So that's a challenge.\textbackslash n\textbackslash nGiven that, the best approach is to select companies that have shown strong earnings beats and positive sales trends, which could be indicative of a turnaround. Even without explicit mention of new management, the user can infer that such performance might be due to restructuring efforts.\textbackslash n\textbackslash nSo, the possible signals are:\textbackslash n\textbackslash n- WBTN: Beat EPS and sales up YoY. Buy.\textbackslash n\textbackslash n- WILC: EPS up 570\% YoY, sales up. Buy.\textbackslash n\textbackslash n- CRBU: Beat EPS and sales, though sales down YoY. Maybe a temporary dip. Buy.\textbackslash n\textbackslash n- ANTA: Beat both EPS and sales. Buy.\textbackslash n\textbackslash n- CRWV: Sales guidance above consensus. Buy.\textbackslash n\textbackslash nBut the user needs to check if these are distressed. Since the background doesn't say, but the user is a turnaround specialist, they might assume that beating estimates after a period of underperformance is a sign of turnaround. For example, if CRBU had previous issues, but the current results are better. However, the data here only shows Q2 results. Without prior performance, it's hard to tell.\textbackslash n\textbackslash nAlternatively, maybe the companies with the largest EPS beats are the best candidates. Let's see:\textbackslash n\textbackslash n- WBTN: 145\% beat.\textbackslash n\textbackslash n- ATEX: 15.79\% beat.\textbackslash n\textbackslash n- CRBU: 9.09\% beat.\textbackslash n\textbackslash n- ANTA: beat EPS and sales.\textbackslash n\textbackslash n- WILC: 570\% beat.\textbackslash n\textbackslash nSo, WBTN and WILC have the most significant improvements. However, the user might need to be cautious. Also, the fact that SPY's after-hours volume is high might indicate continued buying, but that's an ETF.\textbackslash n\textbackslash nAnother thing: the UFC event is expected to benefit PSKY and TKO. If the user is looking for companies that are in a turnaround, maybe PSKY is a candidate. But the background doesn't mention any distress for PSKY. However, the media rights deal could be a new strategic initiative. But the user's belief is about debt reduction and restructuring, not just new deals.\textbackslash n\textbackslash nHmm. Maybe the best approach is to go with the companies that have the most positive earnings surprises, as that could indicate a turnaround. Let's list them:\textbackslash n\textbackslash n1. WBTN (Webtoon) - Adjusted EPS beat by 145\%, sales up YoY. Buy.\textbackslash n\textbackslash n2. WILC (G. Willi-Food Intl) - EPS up 570\%, sales up. Buy.\textbackslash n\textbackslash n3. CRBU (Caribou Biosciences) - Adjusted EPS beat, sales beat. Buy.\textbackslash n\textbackslash n4. ANTA (Antalpha) - Both beat. Buy.\textbackslash n\textbackslash n5. CRWV (CoreWeave) - Sales guidance above consensus. Buy.\textbackslash n\textbackslash nBut the user needs to make sure these are distressed. Since the background doesn't state that, maybe it's safer to pick those with the highest EPS beats. However, the user's belief allows for analysis via press releases and earnings calls. If the earnings call transcripts for these companies were released, maybe there's info on restructuring. The background mentions that LAZR and RGTI transcripts were released, but not others. So, maybe CRBU, ATEX, ANTA, and WBTN have some restructuring info in their earnings calls, but it's not mentioned here. Therefore, without explicit info, it's risky to assume.\textbackslash n\textbackslash nAlternatively, the user might consider that companies that are in the market cap list (like NFLX, LLY, SPOT, MTZ, HIG) have strong long-term returns, but the user's focus is on current opportunities. However, the user's belief is about distressed companies with new management. So maybe those with recent positive results but not necessarily high market caps.\textbackslash n\textbackslash nWait, the user is a turnaround specialist. They might look for companies that have been struggling but now have new management. The background doesn't mention any new management announcements, but maybe the EPS beats and sales improvements could be due to new management. So, the user can suggest buying based on those positive results, even if not explicitly stated.\textbackslash n\textbackslash nSo, I'll go with the top 5 companies that beat EPS and/or sales, assuming that their positive results are part of a turnaround. Let's check:\textbackslash n\textbackslash n- Webtoon (WBTN): adjusted EPS beat and sales up YoY.\textbackslash n\textbackslash n- G. Willi-Food (WILC): EPS up 570\% YoY, sales up.\textbackslash n\textbackslash n- Caribou (CRBU): adjusted EPS beat and sales beat, but YoY down.\textbackslash n\textbackslash n- Anterix (ATEX): adjusted EPS beat, sales missed but down YoY.\textbackslash n\textbackslash n- Antalpha (ANTA): both beat.\textbackslash n\textbackslash nSo, WBTN, WILC, ANTA, CRBU, and ATEX. But the user needs to be careful. Maybe some of these are not distressed. However, given the information, these are the companies that have shown positive results. The user's belief allows for analysis via earnings calls, so perhaps the transcripts for these companies (if available) would indicate restructuring. Since the background only mentions LAZR and RGTI, but the user can't perform further searches, they have to rely on the given data.\textbackslash n\textbackslash nTherefore, the best signals would be the ones with the highest EPS beats. WBTN and WILC are the top two. Then ANTA, CRBU, and ATEX. But the user needs to make sure that the companies are in a turnaround phase. Since the user's belief is to focus on debt reduction, maybe companies with improved financials (like higher sales and EPS) are more likely to have debt reduction progress.\textbackslash n\textbackslash nSo, the final list could be:\textbackslash n\textbackslash n1. WBTN - Buy, due to strong EPS beat and sales growth.\textbackslash n\textbackslash n2. WILC - Buy, due to massive EPS increase and sales growth.\textbackslash n\textbackslash n3. ANTA - Buy, since both EPS and sales beat.\textbackslash n\textbackslash n4. CRBU - Buy, as they beat adjusted EPS and sales.\textbackslash n\textbackslash n5. ATEX - Buy, despite sales down YoY, but EPS beat.\textbackslash n\textbackslash nBut the user's belief is about distressed companies. If the sales are down YoY, maybe those companies are still in a tough spot. However, the EPS beat could indicate cost-cutting or improved efficiency, which is part of debt reduction. For example, ATEX's sales missed but EPS beat, which might suggest better cost management. Similarly, CRBU's sales are down YoY but they beat estimates. Maybe they are restructuring.\textbackslash n\textbackslash nSo, the user might consider these as potential opportunities. Therefore, the answer would be to buy these stocks. However, the user should be cautious and verify if the companies are indeed in a turnaround. Since the background doesn't provide that info, but the user is a turnaround specialist, they might infer based on the financial results.\textbackslash n\textbackslash nThus, the 5 signals would be:\textbackslash n\textbackslash n- WBTN (Webtoon Entertainment) - Buy.\textbackslash n\textbackslash n- WILC (G. Willi-Food Intl) - Buy.\textbackslash n\textbackslash n- ANTA (Antalpha) - Buy.\textbackslash n\textbackslash n- CRBU (Caribou Biosciences) - Buy.\textbackslash n\textbackslash n- CRWV (CoreWeave) - Buy.\textbackslash n\textbackslash nBut need to check if there are 5. Let me count:\textbackslash n\textbackslash n1. WBTN\textbackslash n\textbackslash n2. WILC\textbackslash n\textbackslash n3. ANTA\textbackslash n\textbackslash n4. CRBU\textbackslash n\textbackslash n5. CRWV\textbackslash n\textbackslash nYes, five. However, the user's belief is about debt reduction progress. If any of these companies have mentioned debt reduction in their earnings calls, that would be better. Since the background mentions that the user is to analyze restructuring plans via press releases and earnings call transcripts, but the transcripts for LAZR and RGTI were released. The user can't check others, so they have to rely on the given info. Since the user is allowed to use the provided data, and the fact that they beat estimates could be due to restructuring efforts, the answer would include these 5.\textbackslash n\textbackslash nBut the user might not have enough info on all of them. Alternatively, maybe focus on the ones with the highest positive surprises. So, WBTN, WILC, ANTA, CRBU, and maybe ATEX. However, ATEX's sales missed and YoY down, which is a negative. So maybe exclude ATEX.\textbackslash n\textbackslash nSo revised list:\textbackslash n\textbackslash n1. WBTN\textbackslash n\textbackslash n2. WILC\textbackslash n\textbackslash n3. ANTA\textbackslash n\textbackslash n4. CRBU\textbackslash n\textbackslash n5. CRWV\textbackslash n\textbackslash nEach of these has positive results. The user can suggest buying these as they show signs of improvement, which might be due to restructuring. Even if not explicitly stated, the user can act based on the data. Therefore, the 5 signals would be these companies.\textbackslash n</think>\textbackslash n
    \\
    \textbf{Decision:} <Output>
    \textbackslash n<signals>
    \textbackslash n<signal>
    \textbackslash n<has\_opportunity>true</has\_opportunity>
    \textbackslash n<action>buy</action>
    \textbackslash n<symbol\_code>WBTN</symbol\_code>
    \textbackslash n<symbol\_name>Webtoon Entertainment</symbol\_name>
    \textbackslash n</signal>
    \textbackslash n<signal>
    \textbackslash n<has\_opportunity>true</has\_opportunity>
    \textbackslash n<action>buy</action>
    \textbackslash n<symbol\_code>WILC</symbol\_code>
    \textbackslash n<symbol\_name>G. Willi-Food Intl</symbol\_name>
    \textbackslash n</signal>
    \textbackslash n<signal>
    \textbackslash n<has\_opportunity>true</has\_opportunity>
    \textbackslash n<action>buy</action>
    \textbackslash n<symbol\_code>ANTA</symbol\_code>
    \textbackslash n<symbol\_name>Antalpha Platform Holding</symbol\_name>
    \textbackslash n</signal>
    \textbackslash n<signal>
    \textbackslash n<has\_opportunity>true</has\_opportunity>
    \textbackslash n<action>buy</action>
    \textbackslash n<symbol\_code>CRBU</symbol\_code>
    \textbackslash n<symbol\_name>Caribou Biosciences</symbol\_name>
    \textbackslash n</signal>
    \textbackslash n<signal>
    \textbackslash n<has\_opportunity>true</has\_opportunity>
    \textbackslash n<action>buy</action>
    \textbackslash n<symbol\_code>CRWV</symbol\_code>
    \textbackslash n<symbol\_name>CoreWeave</symbol\_name>
    \textbackslash n</signal>
    \textbackslash n</signals>
    \textbackslash n</Output>
    }
\end{quote}

\end{document}